\documentclass[preprint,number,review,12pt]{elsarticle} 

\usepackage[top=2.5cm,bottom=2.5cm,left=2.5cm,right=2.5cm]{geometry}
\usepackage{amssymb}
\usepackage{graphicx}
\graphicspath{ {./images/} }
\usepackage{multirow} 
\usepackage[centertags]{amsmath} \usepackage{amsfonts}
\usepackage{amsthm}
\usepackage{textcomp}
\usepackage{dblfloatfix}
\usepackage{caption}
\usepackage{cleveref}
\usepackage{subcaption}
\usepackage{lineno}
\DeclareMathAlphabet{\mathpzc}{OT1}{pzc}{m}{it}
\usepackage{amsmath}
\usepackage{bbm}
\usepackage[mathscr]{euscript}
\usepackage{gensymb}
\usepackage{color}
\usepackage{threeparttable}
\usepackage{float} 
\usepackage{blindtext}
\usepackage{multicol}
\usepackage{cuted}
\usepackage{amsmath}
\usepackage{cuted}

\newcommand{\Expect}{{\rm I\kern-.3em E}}

\begin{document}

\begin{frontmatter}




\title{Adaptive Machine Learning-Driven Multi-Fidelity Stratified Sampling for Failure Analysis of Nonlinear Stochastic Systems}

\author[Umich]{Liuyun Xu}
\ead{xliuyun@umich.edu}
\author[Umich]{Seymour M.J. Spence\corref{cor1}}
\ead{smjs@umich.edu}
\cortext[cor1]{Corresponding author}

\address[Umich]{Department of Civil and Environmental Engineering, University of Michigan, Ann Arbor, MI 48109, USA}

\begin{abstract}
Existing variance reduction techniques used in stochastic simulations for rare event analysis still require a substantial number of model evaluations to estimate small failure probabilities. In the context of complex, nonlinear finite element modeling environments, this can become computationally challenging—particularly for systems subjected to stochastic excitation. To address this challenge, a multi-fidelity stratified sampling scheme with adaptive machine learning metamodels is introduced for efficiently propagating uncertainties and estimating small failure probabilities. In this approach, a high-fidelity dataset generated through stratified sampling is used to train a deep learning-based metamodel, which then serves as a cost-effective and highly correlated low-fidelity model. An adaptive training scheme is proposed to balance the trade-off between approximation quality and computational demand associated with the development of the low-fidelity model. By integrating the low-fidelity outputs with additional high-fidelity results, an unbiased estimate of the strata-wise failure probabilities is obtained using a multi-fidelity Monte Carlo framework. The overall probability of failure is then computed using the total probability theorem. Application to a full-scale high-rise steel building subjected to stochastic wind excitation demonstrates that the proposed scheme can accurately estimate exceedance probability curves for nonlinear responses of interest, while achieving significant computational savings compared to single-fidelity variance reduction approaches.
\end{abstract}



\end{frontmatter}
\section{Introduction}
\label{Sec:introduction}
To enable efficient probabilistic analysis---including failure analysis---of structural systems subjected to general stochastic excitations (e.g., seismic or wind loading), considerable advancements have been made in frameworks, modeling techniques, and computational capacity (e.g., \cite{koutsourelakis2004reliability,beck2014optimal,shields2015targeted,melchers2018structural,yi2018xgaussian,arunachalam2022reliability,chuang2022framework,beck2022structural,arunachalam2023generalized,goswami2025neural,deodatis2025spectral,lee2025efficient,giovanis2025accelerating}). Stochastic simulation frameworks that integrate Monte Carlo (MC) methods with high-fidelity modeling environments are commonly employed to propagate uncertainty and estimate failure probabilities for various limit states. While high-fidelity numerical models (e.g., finite element models) can capture essential nonlinear behaviors (e.g., material and geometric nonlinearity), they are often computationally intensive. As a result, when estimating small failure probabilities associated with rare events, these frameworks can become computationally prohibitive due to the large number of model evaluations required.

To alleviate computational demands, variance reduction techniques that maintain accuracy while requiring significantly fewer model evaluations have been explored. Among these approaches, the widely used importance sampling \cite{melchers1989importance} can face challenges when applied to high-dimensional problems encountered in practice, due to the difficulty of identifying a suitable high-dimensional proposal density \cite{au2003important}. Subset Simulation (SuS) and Stratified Sampling (SS), including Generalized Stratified Sampling (GSS) \cite{arunachalam2023generalized}, are capable of estimating small failure probabilities in high-dimensional settings. However, they generally still require several thousand model evaluations to achieve a target level of accuracy \cite{arunachalam2023generalized,arunachalam2023efficient,xu2024collapse}. Compared to the SS approach, SuS can quickly become inefficient when estimating failure probabilities for multiple limit states of interest, as each limit state generally requires an independent implementation of the SuS procedure \cite{au2001estimation}. Alternatively, models with reduced fidelity levels have emerged as a promising solution for approximating system outputs while using significantly lower computational budgets. Low-fidelity models for engineering applications generally fall into two categories: (a) reduced-order models, which simplify the underlying physics by, for example, reducing dynamic degrees of freedom or employing simplified material hysteretic laws \cite{lucia2004reduced,patsialis2020reduced}; and (b) metamodels (or emulators), which provide data-driven approximations of system outputs \cite{li2010evaluation,gidaris2015kriging,li2020efficient}. Recently, artificial intelligence (AI)-driven metamodels have attracted significant research interest due to their potential to achieve orders-of-magnitude speedups. These models are constructed by mapping parameterized input and output spaces using regression or interpolation techniques (e.g., polynomials, neural networks, and Kriging) \cite{gidaris2015kriging,lagaros2012neural,sharma2024physics}. However, it should be noted that low-fidelity models may yield biased or distorted estimates if used directly for uncertainty propagation \cite{li2024multi}. 

To leverage the accuracy of high-fidelity models and the computational efficiency of low-fidelity models, multi-fidelity approaches that integrate outputs from models with varying levels of fidelity have been developed \cite{li2024multi,ng2014multifidelity,peherstorfer2016multifidelity,peherstorfer2018survey}. The core concept is to obtain an accurate estimate by shifting most of the computational burden to the cost-effective, but potentially biased, low-fidelity model evaluations, while applying corrections using a substantially smaller number of high-fidelity model evaluations. Multi-fidelity schemes are generally classified into two categories: (a) approximate control variate approaches \cite{nelson1987control,han2024approximate}, such as Multi-Level Monte Carlo (MLMC) \cite{cliffe2011multilevel,giles2015multilevel} and Multi-Fidelity Monte Carlo (MFMC) \cite{peherstorfer2016optimal,kramer2019multifidelity}; and (b) multi-fidelity surrogate models, such as multi-fidelity Gaussian process models (Cokriging) \cite{yi2021active,renganathan2022multifidelity}. Among these methods, MFMC has gained recognition as an effective means of accelerating standard MC estimation that would otherwise rely entirely on computationally demanding high-fidelity models \cite{peherstorfer2018survey}. Extensive studies have demonstrated the practicality and efficiency of MFMC using various categories of low-fidelity models \cite{peherstorfer2016optimal,patsialis2021multi,jung2024adaptive}. The optimal allocation of evaluations across fidelity levels can be determined by minimizing the variance of the MFMC estimator \cite{peherstorfer2016optimal}. For efficient MFMC implementation, two key characteristics of low-fidelity models are essential: (a) high correlation with the high-fidelity model; and (b) substantial computational savings. However, MFMC can become less effective when addressing small-probability problems. This limitation arises from the need to capture extreme system responses associated with rare failures, which are critical for accurate estimation. Random sampling struggles to generate such rare-event samples efficiently.

This paper develops a Multi-Fidelity Stratified Sampling (MFSS) scheme that employs an adaptive machine learning metamodel as the low-fidelity model to efficiently propagate uncertainties and estimate small failure probabilities. In this approach, a high-fidelity sample set generated through GSS is used to train a deep learning-based low-fidelity model. An adaptive training strategy is proposed to optimize the trade-off between approximation accuracy and computational cost. This strategy aims to minimize the amount of training data while ensuring sufficient unbiased correlation between the high- and low-fidelity models, as determined through K-fold cross-validation. Once developed, a low-fidelity sample set of any size can be rapidly generated. To ensure the estimation remains unbiased, an additional high-fidelity sample set is generated and combined with the low-fidelity sample set. The conditional failure probability for each stratum is then estimated using MFMC with optimally allocated high- and low-fidelity outputs. Subsequently, the overall MFSS estimator is constructed using the total probability theorem. Through application to a full-scale high-rise steel building subjected to stochastic wind excitation, the proposed scheme demonstrates high accuracy and significant computational savings in estimating exceedance probability curves for various system responses. The advantages of the MFSS approach over traditional GSS schemes relying solely on high-fidelity models are further highlighted.


\section{Problem Setting}
\label{Sec:Problem Setting}

Consider a dynamic, nonlinear structural system subjected to stochastic excitation $\boldsymbol{F}(t;\boldsymbol{\theta})$, such as wind or seismic loading, characterized by a vector of uncertain parameters $\boldsymbol{\theta} = \{\theta_{1}, \theta_{2}, \ldots, \theta_{n_\theta}\}^T \in \mathbb{R}^{n_\theta}$, where $n_\theta$ denotes the dimension of $\boldsymbol{\theta}$. In general, the system response, $\boldsymbol{y}(t;\boldsymbol{\theta})$, can be expressed as:
\begin{equation}
\boldsymbol{y}(t;\boldsymbol{\theta}) = \mathcal{M}(\boldsymbol{F}(t;\boldsymbol{\theta}))
\label{eq: eom}
\end{equation}
where $\mathcal{M}(\cdot)$ represents a generally high-dimensional and computationally intensive nonlinear finite element model (hereafter referred to as the ``high-fidelity model'') that maps the stochastic excitation to the structural response (e.g., displacements at degrees of freedom). The quantity of interest, $Z$, is derived from the time-dependent system response $\boldsymbol{y}(t;\boldsymbol{\theta})$ through a function $f$, i.e., $Z = f(\boldsymbol{y}(t;\boldsymbol{\theta}))$. This function may involve various operations, such as extracting the peak displacement across all degrees of freedom. Let $p(\boldsymbol{\theta})$ denote the probability density function of $\boldsymbol{\theta}$. The problem of interest is to estimate the probability that $Z$ exceeds a critical threshold (limit state) $z_i$, denoted as $P_{fi} = P(Z > z_i)$. This probability can be expressed as the expected value of a consequence measure $h_i$, defined as a function of $\boldsymbol{\theta}$:
\begin{equation}
P_{fi} = \mathbb{E}[h_i(\boldsymbol{\theta})] = \int_{\Theta} h_i(\boldsymbol{\theta}) \, p(\boldsymbol{\theta}) \, d\boldsymbol{\theta}
\label{eq: pf}
\end{equation}
where the subscript $i$ refers to the $i$th limit state of interest, and $\mathbb{E}[\cdot]$ denotes the expectation operator. The function $h_i(\cdot)$ can take various forms, including: (a) an indicator function, $h_i(\cdot) = \mathbb{I}(Z > z_i)$, which equals 1 when $Z > z_i$ and 0 otherwise~\cite{patsialis2021multi}; or (b) a log-transformed kernel estimator, $h_i(\cdot) = 1 - K\left( \frac{\ln(z_i / Z)}{b} \right)$, where $K(\cdot)$ is a distribution function with a positive kernel (e.g., the standard normal distribution~\cite{simonoff2012smoothing,suksuwan2018optimization}), and $b$ is a bandwidth parameter~\cite{silverman2018density}.

Accurately estimating small failure probabilities associated with rare events using Eq.~\eqref{eq: pf} typically requires a large number of high-fidelity model evaluations (often on the order of thousands), leading to significant computational challenges. To address this, this paper develops an MFSS framework that implements MFMC within a stratified probability space to enable efficient estimation of small failure probabilities. To further enhance computational efficiency, a deep learning-based metamodel is constructed and employed as the low-fidelity model within the proposed MFSS framework.


\section{Background Schemes}
\subsection{Generalized Stratified Sampling}
\label{subSec: Generalized Stratified Sampling}

Stratified sampling has been demonstrated as an efficient method for estimating small failure probabilities, enabling significant variance reduction compared to the direct MC approach \cite{arunachalam2023efficient}. In this scheme, the sample probability space is partitioned into $N_s$ mutually exclusive and collectively exhaustive subevents, $E^k$, for $k=1,2,...,N_s$, termed strata. This allows for samples to be drawn from each stratum, including those associated with extreme responses related to rare events. To address problems where stratification based on the basic random variables does not yield obvious computational benefits—such as when no single random variable dominates the response—\citet{arunachalam2023generalized} proposed generalized stratified sampling, GSS. This approach focuses on partitioning the probability space with respect to an intermediate quantity that is highly correlated with the system response, referred to as the stratification variable ($SV$). To ensure computational efficiency of GSS, the cost of evaluating $SV$ should be significantly less than the cost of directly evaluating the limit state function.

GSS adopts a double sampling approach. In Phase I, direct MC techniques are employed to explore the probability space of $SV$ by employing a large number of MC samples, $\hat{N}_{MC}$. By leveraging the principles of random sampling, the strata probabilities can be estimated as $P(E^k)\approx \hat{N}^k_{MC}/\hat{N}_{MC}$ for $k=1,..., N_s$, where $\hat{N}_{MC}^k$ represents the number of samples out of $\hat{N}_{MC}$ lying in the $k$th stratum, denoted as strata-wise samples. To ensure an adequate number of samples in each stratum, a large number of MC samples should be generated. Specifically, approximately $\hat{N}_{MC}=10^{m+2}$ evaluations of $SV$ are required to yield an estimation of $P(E^k)\approx 10^{-m}$ with a coefficient of variation (COV) of 10\%, resulting in roughly $10^2$ samples within the stratum $E^k$ \cite{lelievre2018ak}. In Phase II, from the $\hat{N}_{MC}^k$ strata-wise samples, $N_{MC}^k$ samples are selected to perform limit state evaluations and estimate the conditional failure probability, $P_{fi}^k=P(Z>z_i|E^k)$,  for each stratum. Therefore, the total failure probability can be estimated as:
\begin{align}
& P_{fi} =\sum_{k=1}^{N_s} P_{fi}^k \cdot P(E^k) \nonumber\\
&\approx \sum_{k=1}^{N_s}\left(\frac{\sum_{j=1}^{N_{MC}^k}h_i(\boldsymbol{\theta}^k_j)}{N_{MC}^k}\right) \cdot  P(E^k) =\sum_{k=1}^{N_s}\hat{s}_{i,k} \cdot P(E^k) =\hat{H}_{i,GSS}
\label{eq: generalized ss}
\end{align}
where $\boldsymbol{\theta}^k_j$ is the $j$th selected realization of $\boldsymbol{\theta}$ in stratum $k$, and $\hat{s}_{i,k}$ denotes the MC estimator of the stratum-wise conditional failure probability associated with the $i$th limit state. Another important property of this scheme is the estimator variance, which can be written as follows \cite{arunachalam2023generalized}:
\begin{align}  
\mathbb{V}\left[\hat{H}_{i,GSS}\right] 
=\mathbb{V}\left[
\frac{\sum_{j=1}^{\hat{N}_{MC}}h_i(\boldsymbol{\theta}_j)}{\hat{N}_{MC}}\right]
+ \sum_{k=1}^{N_s} \left[P(E^k)\right]^2 \cdot \frac{\mathbb{V}\left[h_i(\boldsymbol{\theta}^k)\right]}{N^k_{MC}}
\cdot \left(1-\nu_k\right) 
\label{eq: stratified-var} 
\end{align} 
where $\mathbb{V}[\cdot]$ is the variance operator, $\nu_k=N_{MC}^k/\hat{N}_{MC}^k\in (0,1]$ represents the proportion of samples in the $k$th stratum from Phase I considered in Phase II for failure probability evaluations. Detailed derivations of Eq.~\eqref{eq: stratified-var} can be found in \cite{arunachalam2023generalized}.

If a large number of MC samples is used to estimate the strata probabilities ($\hat{N}_{MC}\to \infty$), $P_{E^k}$ will tend toward its true population value. In this context, the number of samples used to evaluate the limit state function is much fewer relative to the total number of strata-wise samples ($\nu_k \to 0$). As a result, GSS tends towards classic SS applied over a known probability space of the $SV$. Under these conditions, the variance of the estimator in Eq.~\eqref{eq: stratified-var} can be simplified as:
\begin{align}   \mathbb{V}\left[\hat{H}_{i,GSS}\right] 
=\sum_{k=1}^{N_s} \left[P(E^k)\right]^2 \cdot \frac{\mathbb{V}\left[h_i(\boldsymbol{\theta}^k)\right]}{N^k_{MC}} 
 =\sum_{k=1}^{N_s} \left[P(E^k)\right]^2 \cdot \mathbb{V}\left[\hat{s}_{i,k}\right]
\label{eq: stratified-var-simplify} 
\end{align}
The above assumption is feasible as long as $SV$ is cheap to evaluate. 

Finally, it can be shown that the COV of the GSS estimator, $\kappa_{i,GSS}$, can be estimated as:
\begin{align}   \kappa_{i,GSS}=\frac{\sqrt{\mathbb{V}\left[\hat{H}_{i,GSS}\right]}}{\hat{H}_{i,GSS}}
= \frac{
\sqrt{\sum_{k=1}^{N_s} \left[P(E^k)\right]^2 
\cdot \mathbb{V}\left[\hat{s}_{i,k}\right] }}
{\sum_{k=1}^{N_s} \hat{s}_{i,k} \cdot P(E^k) } 
\label{eq: stratified-cov-simplify} 
\end{align}


\subsection{Multi-Fidelity Monte Carlo}
\label{subSec: MFMC}

A MFMC scheme that effectively integrates high- and low-fidelity model outputs can provide an unbiased estimator with substantial variance reduction \cite{patsialis2021multi}. Generally, a MFMC scheme can incorporate a range of numerical models with varying fidelity levels. This work focuses on a bi-fidelity setting, utilizing a single low-fidelity model alongside a high-fidelity model. For clarity, the terms $HF$ and $LF$ will be used to denote the high- and low-fidelity models. The respective computational costs of the $HF$ and $LF$ models are denoted by $c_{HF}$ and $c_{LF}$, where ideally $c_{HF} \gg c_{LF}$. The MFMC estimator, $\hat{H}_{i,MF}$, for Eq. \eqref{eq: pf} can be mathematically expressed as \cite{peherstorfer2016optimal,patsialis2021multi}:
\begin{align}     
P_{fi} \approx  \hat{H}_{i,MF} & =\hat{s}_{HF}^{N_{i,HF}}+a_i\left(\hat{s}_{LF}^{N_{i,LF}}-\hat{s}_{LF}^{N_{i,HF}}\right) \nonumber \\
& = \frac{1}{N_{i,HF}}\sum_{j=1}^{N_{i,HF}}h_{i,HF}(\boldsymbol{\theta}_{j}) \nonumber \\
& +a_i\left(\frac{1}{N_{i,LF}}\sum_{j=1}^{N_{i,LF}}h_{i,LF}(\boldsymbol{\theta}_{j})-\frac{1}{N_{i,HF}}\sum_{j=1}^{N_{i,HF}}h_{i,LF}(\boldsymbol{\theta}_{j})
\right) 
\label{eq: MFMC}
\end{align}  
where $\hat{s}_*^l$ represents the MC estimator using $l$ evaluations of the $*$ model (e.g., $HF$ or $LF$); $N_{i,HF}$ and $N_{i,LF}$ are the number of $HF$ and $LF$ samples used for evaluating the $i$th limit state of interest; $h_{i,HF}(\cdot)$ and $h_{i,LF}(\cdot)$ represent consequence measures associated with the $i$th limit state of interest based on the high- and low-fidelity model evaluations; $a_i$ is the control variate coefficient. In Eq.~\eqref{eq: MFMC}, $\hat{s}_{LF}^{N_{i,HF}}$ reuses the first $N_{i,HF}$ model evaluations that are also used for $\hat{s}_{LF}^{N_{i,LF}}$, making the two estimators dependent~\cite{peherstorfer2016optimal}. Nevertheless, the unbiasedness of the estimator $\hat{H}_{i,MF}$ holds, provided that $\hat{s}_{LF}^{N_{i,LF}}$ and $\hat{s}_{LF}^{N_{i,HF}}$ share the same expectation, which follows from the unbiasedness of the MC estimator. A proof of this result is provided in \ref{appen: MFMC}.

To enhance the efficiency of the MFMC estimator, $a_i$, as well as the sample allocation ratio, $r_i={N_{i,LF}}/{N_{i,HF}}$, can be optimally determined through minimizing the estimator variance leading to
\cite{peherstorfer2016optimal,patsialis2021multi}:
\begin{equation}
    a_i^*=\rho_i \cdot \sqrt{\frac{\mathbb{V}\left[h_{i,HF}\right]}{\mathbb{V}\left[h_{i,LF}\right]}}
    \label{eq: a}
\end{equation}
\begin{equation}
    r_i^*=
\frac{N_{i,LF}}{N_{i,HF}}=\sqrt{\frac{c_{HF}\cdot \rho_i^2}{c_{LF} \cdot (1-\rho_i^2)}}
    \label{eq: r}
\end{equation}
where $\rho_i$ denotes the correlation coefficient  between $h_{i,HF}$ and $h_{i,LF}$. The minimized MFMC estimator variance can be expressed as:
\begin{equation}
\mathbb{V}\left[\hat{H}_{i,MF}\right]=\frac{\mathbb{V}[h_{i,HF}(\boldsymbol{\theta})]}{N_{i,HF}}\cdot \left(1-\left(1-\frac{1}{r_i^*}\right)\cdot \rho_i^2 \right)
\label{eq: MFMCvar}
\end{equation}
From Eq. (\ref{eq: MFMCvar}), it can be demonstrated that when an uncorrelated $LF$ model is considered ($\rho_i\approx 0$), the MFMC estimator essentially reduces to that of direct MC using $N_{i,HF}$ $HF$ samples. Conversely, when incorporating a perfect $LF$ model ($\rho_i\approx 1$), MFMC  predominantly relies on $LF$ model evaluations ($ r_i^* \to \infty $). It is evident that increasing $\rho_i$ and $c_{HF}/c_{LF}$ results in an increase in $r_i^*$. This illustrates how the MFMC scheme achieves estimation precision with computational savings by shifting evaluations onto a highly correlated and cost-effective $LF$ model. As shown in~\cite{patsialis2021multi}, achieving the same estimator variance using direct MC simulation based on $HF$ model outputs requires the following number of model evaluations:
\begin{equation}
N_{i,\text{sim}} = N_{i,HF} \cdot \left(1 - \left(1 - \frac{1}{r^*_i}\right) \cdot \rho_i^2 \right)^{-1}
\label{eq: N mc}
\end{equation}
where $N_{i,\text{sim}}$ is the number of required high-fidelity model evaluations.


\section{Proposed Approach}
\subsection{Multi-Fidelity Generalized Stratified Sampling}
\label{SubSec: MFSS}

To further enhance computational efficiency in estimating small failure probabilities, this work proposes a scheme that performs MFMC within a stratified probability space, termed multi-fidelity stratified sampling, or MFSS. Consistent with GSS, as outlined in Sec.~\ref{subSec: Generalized Stratified Sampling}, the scheme consists of two implementation phases. In Phase I, direct MC sampling is employed to estimate the probability distribution of a predefined $SV$ by generating a large number, $\hat{N}_{MC}$, of MC samples. Through application of the approach outlined in \citet{arunachalam2023generalized}, $\hat{N}_{MC}^k$ strata-wise samples can be collected for each stratum, enabling the estimation of strata probabilities $P(E^k)$. In Phase~II, the conditional failure probability within each stratum is approximated using MFMC, as described in Sec.~\ref{subSec: MFMC}. The proposed MFSS approach is designed to efficiently combine the benefits of GSS and MFMC, significantly reducing computational cost while maintaining accuracy in the estimation of small failure probabilities.

In the MFSS scheme, a $HF$ sample set consisting of $N_{\text{train}}$ random samples from each stratum, selected from the $\hat{N}_{MC}^k$ strata-wise samples, is used as training data to develop a deep learning-based metamodel, which serves as the $LF$ model (details on this model are provided in Sec. \ref{subSec: Adaptive LF Model}). This $LF$ model is both computationally efficient and well correlated with the $HF$ model, enabling efficient MFSS implementation. Once developed, strata-wise $HF$ and $LF$ outputs are combined to estimate the conditional failure probability, $P_{fi}^k$, for each stratum using MFMC, as follows:
\begin{align}
P_{fi}^k\approx \hat{H}_{i,MF}^{k}=
    & \frac{1}{N_{i,HF}^k}\sum_{j=1}^{N_{i,HF}^k}h_{i,HF}(\boldsymbol{\theta}_{j}^k)  \nonumber \\
   & +a_i^k\left(\frac{1}{N_{i,LF}^k}\sum_{j=1}^{N_{i,LF}^k}h_{i,LF}(\boldsymbol{\theta}_{j}^k)-\frac{1}{N_{i,HF}^k}\sum_{j=1}^{N_{i,HF}^k}h_{i,LF}(\boldsymbol{\theta}_{j}^k)\right)
    \label{eq: mfss-k-no assumption}
\end{align}
where $N_{i,HF}^k$ and $N_{i,LF}^k$ are the number of $HF$ and $LF$ samples generated for evaluating the $i$th limit state within the $k$th stratum, and $a_i^k$ is the control variate coefficient of the estimator for the $k$th stratum associated with the $i$th limit state. To ensure the unbiasness of the MFMC estimator, the $LF$ outputs are generated using samples that were not used during training.
This guarantees that the expected values of the MC estimators based on ${N_{i,LF}^k}$ and ${N_{i,HF}^k}$ $LF$ evaluations, namely $\hat{s}_{LF}^{N_{i,LF}^k} = \frac{1}{N_{i,LF}^k} \sum_{j=1}^{N_{i,LF}^k} h_{i,LF}(\boldsymbol{\theta}_{j}^k)$ and $\hat{s}_{LF}^{N_{i,HF}^k} = \frac{1}{N_{i,HF}^k} \sum_{j=1}^{N_{i,HF}^k} h_{i,LF}(\boldsymbol{\theta}_{j}^k)$, are equal, which is a prerequisite for maintaining the unbiasedness of the overall estimation.

To simplify MFMC implementation for multiple limit states of interest, the correlation coefficient between $h_{i,HF}$ and $h_{i,LF}$ is approximated by the correlation between appropriate $HF$ and $LF$ model outputs computed in a reduced space and aggregated across all strata. This correlation coefficient, denoted as $\rho$, will be discussed thoroughly in Sec. \ref{ssubsec: adaptive training}. This approximation overcomes a common challenge in MFMC, which involves performing varying numbers of model evaluations when estimating the failure probabilities for multiple limit states of interest. Indeed, within this setting, the ratio of $LF$ to $HF$ model evaluations associated with each stratum remains constant across the various limit states of interest. In addition, because $\rho$ is aggregated across all strata, the ratio is also independent of the particular stratum. Following the discussion in Sec.~\ref{subSec: MFMC}, the optimal value of this ratio can be expressed as:
\begin{equation}
    r^*=N_{LF}^k/ N_{HF}^k= \sqrt{\frac{c_{HF}\cdot \rho^2}{c_{LF} \cdot (1-\rho^2)}}
    \label{eq: r_critical}
\end{equation}
where $N_{HF}^k$ and $N_{LF}^k$ denote the number of $HF$ and $LF$ samples used in each stratum for evaluating the limit states of interest.

In this work, an equal allocation across all strata is utilized, that is, $N_{HF}^k=N_{HF}$ and $N_{LF}^k=N_{LF}$, for $k=1,2,..., N_s$. With this setup, the MFMC estimator in Eq.~\eqref{eq: mfss-k-no assumption} can be simplified as:
\begin{align}
\hat{H}_{i,MF}^{k}=
    & \frac{1}{N_{HF}}\sum_{j=1}^{N_{HF}}h_{i,HF}(\boldsymbol{\theta}_{j}^k)  \nonumber \\
   & +a_i^k\left(\frac{1}{N_{LF}}\sum_{j=1}^{N_{LF}}h_{i,LF}(\boldsymbol{\theta}_{j}^k)-\frac{1}{N_{HF}}\sum_{j=1}^{N_{HF}}h_{i,LF}(\boldsymbol{\theta}_{j}^k)\right)
    \label{eq: mfss-k}
\end{align}
where $a_i^k$ can be optimally determined as: 
\begin{equation}
  \left(a_i^k\right)^*=\rho \cdot \sqrt{\mathbb{V}\left[h_{i,HF}^k\right] / \mathbb{V}\left[h_{i,LF}^k\right]}
  \label{eq: a_critical}
\end{equation}
From the total probability theorem, it follows that the overall MFSS estimator, $\hat{H}_{i,MS}$, can be expressed as:
\begin{equation}
P_{fi}=\sum_{k=1}^{N_s} P_{fi}^k \cdot P(E^k)\approx \sum_{k=1}^{N_s} \hat{H}_{i,MF}^{k} \cdot P(E^k) =\hat{H}_{i,MS}
    \label{eq: mfss}
\end{equation}

Notably, the number of MC samples used in Phase-I sampling, $\hat{N}_{MC}$, is recommended to be large in this approach. This ensures not only sufficient strata-wise samples for both $HF$ and $LF$ evaluations but also an unbiased estimation of the strata probabilities. In this context, the variance of the MFSS estimator can be expressed as follows:
\begin{align}
\mathbb{V}[\hat{H}_{i,MS}]=
& \sum_{k=1}^{N_s} P(E^k)^2 \cdot \mathbb{V} \left[\hat{H}_{i,MF}^{k}\right]
\nonumber\\
&= \sum_{k=1}^{N_s}  P(E^k)^2 \cdot \frac{\mathbb{V}\left[h_{i,HF}(\boldsymbol{\theta}^k)\right]}{N_{HF}} \cdot
\left(1-\left(1-\frac{1}{r^*}\right)\cdot \rho^2 \right)
    \label{eq: mfss-var}
\end{align}
Subsequently, the COV of the MFSS estimator associated with the $i$th limit state, can be defined as $\kappa_{i,MS}=\sqrt{\mathbb{V}[\hat{H}_{i,MS}]}/\hat{H}_{i,MS}$. To achieve the same variance, $HF$-based GSS would require the following number of model evaluations from each stratum:
\begin{equation}
N_{GSS}=N_{HF} \cdot \left(1-\left(1-\frac{1}{r^*}\right)\cdot \rho^2 \right)^{-1}
    \label{eq: N_ss}
\end{equation}
To assess the efficiency of the proposed MFSS, the computational speed-up, $sp_{MS}$, relative to GSS based solely on $HF$ model outputs with equivalent accuracy, can be expressed as:
\begin{align}
  sp_{MS}&= \frac{c_{HF}\cdot N_{GSS}}{c_{HF}\cdot \left(N_{HF} + N_{\text{train}}\right)+c_{LF}\cdot N_{LF}} \nonumber\\
  &= \frac{ N_{GSS}}{N_{HF} + N_{\text{train}}+\frac{1}{c_{HF}/c_{LF}}\cdot N_{LF}} 
\label{eq: sp_mfss}
\end{align}
where $N_{\text{train}}$ is the total number of $HF$ training samples used to calibrate the $LF$ model in each stratum. It is important to note that Eq.~\eqref{eq: sp_mfss} holds for any limit state of interest. 

Compared to the $HF$-based GSS scheme, the proposed MFSS approach offers significant computational savings without compromising accuracy by efficiently integrating strata-wise $HF$ and $LF$ model outputs. Furthermore, conventional MFMC inherently relies on random sampling and lacks a systematic mechanism to effectively capture rare events, limiting its efficiency for estimating small failure probabilities. By introducing stratification within the MFSS framework, the proposed approach explicitly targets the tails of the distribution, significantly enhancing the representation of extreme samples. As a result, MFSS extends the applicability of MFMC to rare event estimation, achieving significant computational efficiency without sacrificing accuracy. This is further supported by the development of an effective $LF$ model through the combination of stratified sampling and deep learning techniques.


\subsection{Adaptive Metamodel Development}
\label{subSec: Adaptive LF Model}
\subsubsection{Preamble}
\label{ssubSec:Preamble}

Within the MFSS scheme, a deep learning-based metamodel is developed using $HF$ model evaluations for training, serving as the $LF$ model. Deep neural networks are employed due to their ability to capture complex nonlinear relationships while offering substantial computational efficiency (i.e., over three orders of magnitude faster than direct computations using the full $HF$ model)~\cite{li2022metamodeling}. The stochastic excitation, $\boldsymbol{F}(t;\boldsymbol{\theta})$, which captures phenomena such as record-to-record variability in seismic applications, serves as the source of input uncertainties. For simplicity, the input stochastic excitation, $\boldsymbol{F}(t;\boldsymbol{\theta})$, and the output system response, $\boldsymbol{y}(t;\boldsymbol{\theta})$, will hereafter be denoted by $\boldsymbol{F}(t)$ and $\boldsymbol{y}(t)$, respectively. Generally, $\boldsymbol{F}(t)$ and $\boldsymbol{y}(t)$, representing the $n$-dimensional excitation and system response (i.e., $\boldsymbol{F}(t) = \{F_1(t), ..., F_n(t)\}^T$ and $\boldsymbol{y}(t) = \{y_1(t), ..., y_n(t)\}^T$), are discretized into $t_n$ time steps. The discretized representations of $\boldsymbol{F}(t)$ and $\boldsymbol{y}(t)$ are denoted as $\tilde{\boldsymbol{F}}(t_i) = \{F_1(t_i), ..., F_n(t_i)\}^T$ and $\tilde{\boldsymbol{y}}(t_i) = \{y_1(t_i), ..., y_n(t_i)\}^T$ for $i = 1, ..., t_n$. Consequently, the $LF$ model development focuses on metamodeling the sequence-to-sequence mapping from the discretized stochastic excitation, $\tilde{\boldsymbol{F}}$, to the discretized system response, $\tilde{\boldsymbol{y}}$.


\subsubsection{Reduced Space}
\label{ssubsec:ReducedS}
Directly creating neural networks mapping from $\tilde{\boldsymbol{F}}(t_i)$ to $\tilde{\boldsymbol{y}}(t_i)$ for practical engineering systems, which often involve high-dimensional input and output spaces ($n$ is often on the order of thousands in real-world applications), can be both computationally prohibitive and numerically unstable. To address these challenges, effective dimensionality reduction techniques have been extensively investigated~\cite{patsialis2020reduced,bamer2017new,Li2021Response}. This work adopts a Proper Orthogonal Decomposition (POD)-based model order reduction \cite{li2022metamodeling,Li2021Response,kerschen2002physical,volkwein2013proper}. In this approach, the $n$-dimensional discretized system output, $\tilde{\boldsymbol{y}} \in \mathcal{R}^{n}$ can be approximately expressed as $\tilde{\boldsymbol{y}}\approx \boldsymbol{\Phi} \tilde{\boldsymbol{q}}$, where $\tilde{\boldsymbol{q}} \in \mathcal{R}^{n_r}$ collects the discretized reduced outputs ($n_r$ is the reduced dimensionality with $n_r \ll n$), and $\boldsymbol{\Phi}$ is the transformation matrix defining the projection into the reduced space. By carrying out a singular value decomposition (SVD) on a matrix $\boldsymbol{X}\in \mathcal{R}^{n\times n_t}$, which is constructed by collecting $n_t$ snapshots from the discretized system outputs across a set of training samples, the dimensions of the reduced space, $n_r$, can be determined by:
\begin{equation}
    \frac{\sum_{l=1}^{n_r} \lambda_l^2}{\sum_{l=1}^{n} \lambda_l^2 } \geq \eta
    \label{eq: eta}
\end{equation}
where $\lambda_l$ is the $l$th largest singular value of $\boldsymbol{X}$ and $\eta \in (0,1]$ defines a truncation threshold that reflects a trade-off between accuracy and efficiency. The transformation matrix $\boldsymbol{\Phi}$ defining the reduced space can be constructed through collecting the first $n_r$ left singular vectors, termed POD modes. As $\eta$ increases and more POD modes are included, the accuracy improves, but the dimensionality of the reduced space also increases. The input projection is defined using the transpose of the output reduction basis, $\boldsymbol{\Phi}^T$, i.e., $\tilde{\boldsymbol{p}} = \boldsymbol{\Phi}^T\tilde{\boldsymbol{F}}$, ensuring that both inputs and outputs are consistently expressed within the same reduced subspace. Implementing this reduction converts the original high-dimensional mapping $\tilde{\boldsymbol{F}} \rightarrow \tilde{\boldsymbol{y}}$ in the physical space to a significantly lower-dimensional mapping $\tilde{\boldsymbol{p}} \rightarrow \tilde{\boldsymbol{q}}$ in the reduced space, thereby not only improving training efficiency but also enhancing the model’s ability to capture and generalize complex system behavior by isolating the dominant response modes that govern the system’s dynamics. 


\subsubsection{Deep Learning-Based Metamodeling}
\label{ssubsec: Deep learning-based metamodel}

To establish the aforementioned mapping in the reduced space, several studies have explored Long Term Short Memory (LSTM) networks for their effectiveness with sequential data such as discrete time-series data \cite{li2022metamodeling,zhang2020physics,li2024deep,atila2025metamodeling}. LSTM networks have been demonstrated to outperform traditional Recurrent Neural Networks, which are prone to gradient vanishing or exploding problems, particularly when addressing long-term dependencies. More recently, GRU-based networks have been introduced in the application of nonlinear dynamic system response prediction \cite{wu2023prediction,gao2025dynamic}. GRU units simplify the architecture of LSTM networks by using fewer trainable parameters, replacing the input, output, and forget gates of LSTM networks with just two gates: an update gate and a reset gate \cite{shewalkar2019performance,nosouhian2021review}. 

\begin{figure*}[t]
  \centering
  \includegraphics[scale=0.4]{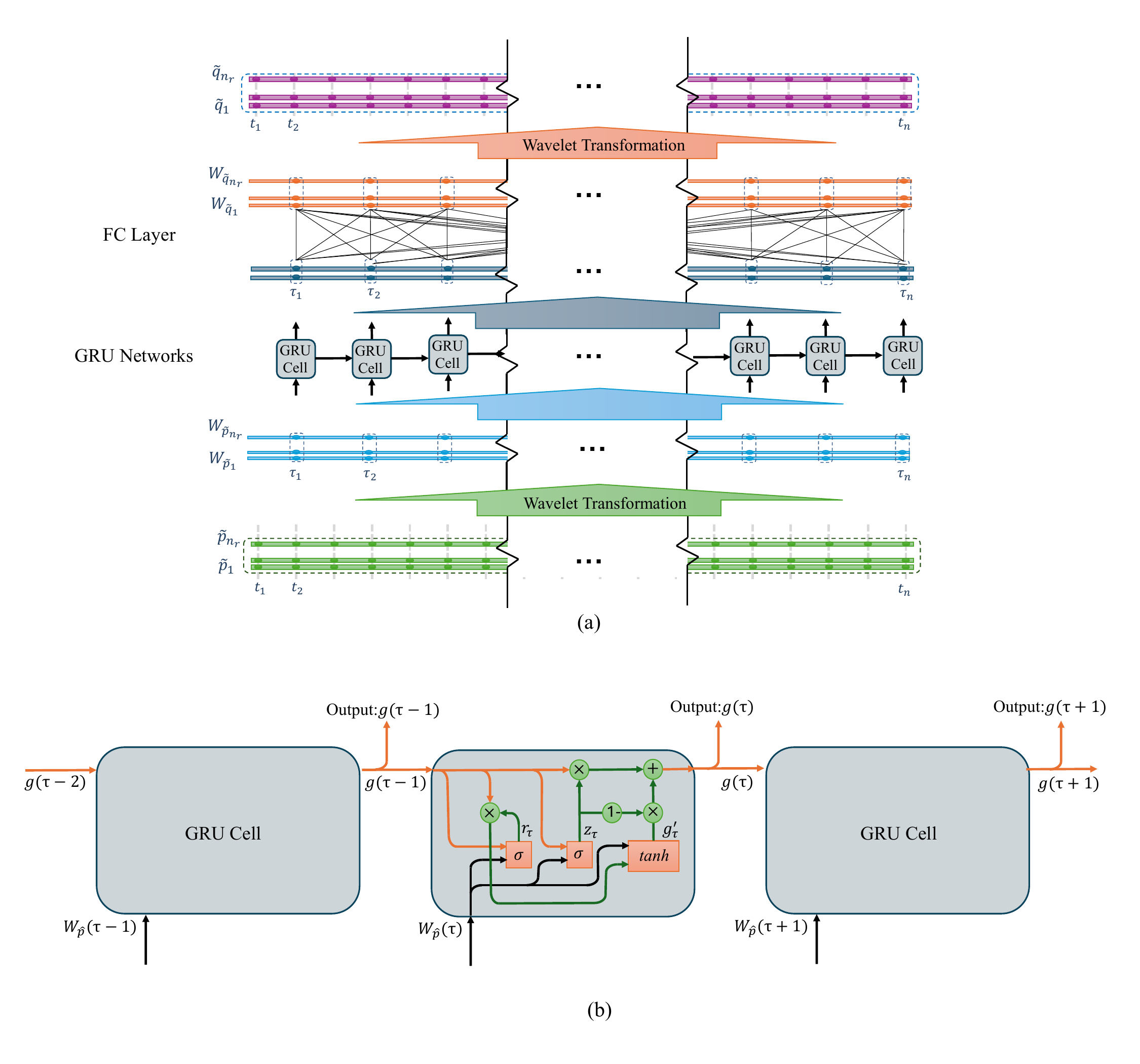}
  \caption{GRU-based metamodeling framework in the reduced space: (a) Overall architecture; (b) Typical GRU cell structure.}
  \label{fig: metamodel}
\end{figure*}

The neural network architecture developed in this work for representing the reduced space mapping $\tilde{\boldsymbol{p}} \rightarrow \tilde{\boldsymbol{q}}$ involves GRU layers, paired with a dropout layer, added immediately after each GRU layer for overfitting mitigation \cite{srivastava2014dropout}. Another benefit of incorporating a dropout layer is the potential to speed up the training process, as fewer parameters remain in the network after dropout. A Fully Connected (FC) layer is appended after the final GRU layer to provide additional flexibility in learning the transformation between the GRU outputs and the final predicted response. Additionally, when dealing with sequence-to-sequence mapping involving a large number of discrete time steps, this setup will include a correspondingly large number of GRU cells, potentially leading to substantial computational demand and computer memory requirements. To address this issue, a Daubechies wavelet-based approximation \cite{cohen1992biorthogonal} is carried out prior to training to reduce the sequence length from $t_n$ to $\tau_n$, thereby simplifying the input–output mapping \cite{li2022metamodeling,le2015reduced,wang2020knowledge}. Consequently, the GRU-based metamodeling framework is centered on learning the mapping between discrete sequences of input wavelet coefficients, $\boldsymbol{W}_{\tilde{\boldsymbol{p}}} = \{W_{\tilde{p}_1},...,W_{\tilde{p}_{n_r}} \}^T$, to discrete sequences of output wavelet coefficients, $\boldsymbol{W}_{\tilde{\boldsymbol{q}}} =\{W_{\tilde{q}_1},...,W_{\tilde{q}_{n_r}} \}^T$. Fig.~\ref{fig: metamodel} illustrates the GRU-based metamodeling framework. The reduced inputs $\tilde{\boldsymbol{p}}$ are first processed by wavelet transformation to reduce the sequence length. Subsequently, the GRU networks, coupled with the FC layer, establish the sequence-to-sequence mapping from input wavelet coefficients, $\boldsymbol{W}_{\tilde{\boldsymbol{p}}}$, to the output wavelet coefficients, $\boldsymbol{W}_{\tilde{\boldsymbol{q}}}$. These output wavelet coefficients are then transformed to the reduced outputs, $\tilde{\boldsymbol{q}}$.

\subsubsection{Adaptive Training Scheme}
\label{ssubsec: adaptive training}

A key requirement for the efficient implementation of MFSS is that the $LF$ model be sufficiently correlated with the $HF$ model. To quantify this correlation, the following reduced space weighted correlation coefficient is proposed:
 \begin{equation}
\rho_v=\frac{\sum_{l=1}^{n_r} \lambda_l \cdot \rho_l}{\sum_{l=1}^{n_r} \lambda_l}
      \label{eq: rho}
 \end{equation}
where $\rho_l$ denotes the correlation coefficient between a reduced output of interest (e.g., peak absolute reduced displacement $\hat{q}_l=\text{max}[|q_l(t)|])$ associated with the $HF$ and $LF$ models and the $l$th mode, while the $l$th largest singular value of the reduced space, $\lambda_l$, acts as a weighting factor. Defining $\rho_v$ in the reduced space through Eq.~(\ref{eq: rho}) provides a single, aggregated measure of correlation, thereby eliminating the need to compute separate correlation coefficients for each quantity of interest in the full physical space. To obtain an unbiased estimate of $\rho_v$ while reducing computational cost, K-fold cross-validation can be employed. In this approach, the $HF$ samples used to train the $LF$ model are partitioned into $k$ equally sized folds. In each round, one fold is held out for testing while the remaining $k{-}1$ folds are used for training. The model correlation coefficient, $\rho_v$, is then evaluated on the held-out fold to reflect the performance of the $LF$ model on unseen data. This process is repeated across all $k$ folds, and the mean correlation coefficient, $\bar{\rho}_v$, is computed by averaging the resulting values of $\rho_v$. This cross-validation strategy mitigates optimistic bias that may arise when evaluating $\rho_v$ on the training data alone, thereby providing a more robust and generalizable estimate. The corresponding coefficient of variation, $\delta_v$, is computed to quantify the dispersion of the estimated correlation across folds. Good results are typically obtained using 5- to 10-fold partitions \cite{fushiki2011estimation}. The resulting mean correlation coefficient, $\bar{\rho}_v$, is then adopted in the MFSS framework (i.e., $\rho = \bar{\rho}_v$) to determine the optimal control variate coefficients and guide the allocation of $HF$ and $LF$ samples.

In theory, a $LF$ model that is perfectly correlated with the $HF$ model—yielding $\bar{\rho}_v = 1$ and $\delta_v = 0$—can be achieved by continuously increasing the amount of training data and appropriately tuning the complexity of the neural network architecture. However, this approach imposes substantial computational costs. In practice, within the MFSS framework, it is not necessary for the $LF$ model to match the accuracy of the $HF$ model. Rather, the $HF$ model evaluations are used to ensure the accuracy guarantees of the multi-fidelity estimator, even when the $LF$ model provides a relatively coarse approximation of the $HF$ outputs \cite{peherstorfer2018survey,peherstorfer2019multifidelity}. Therefore, a cost-effective metamodel that is sufficiently correlated with the $HF$ model is recommended as the $LF$ model within the MFSS framework. To construct such a model, an adaptive strategy is employed to seek a quasi-optimal trade-off between approximation accuracy and computational efficiency. The approach begins by training the $LF$ model on a small dataset and incrementally adds a fixed number of samples in each iteration until a target correlation, $\bar{\rho}_v^*$, and COV, $\delta_v^*$, are met. The objective is to minimize the required training data while ensuring the $LF$ model achieves a target correlation with the $HF$ model.


\subsection{Overall Framework}
\label{subSec:Overall Framework}

\begin{figure*}[]
  \centering
  \includegraphics[scale=0.47]{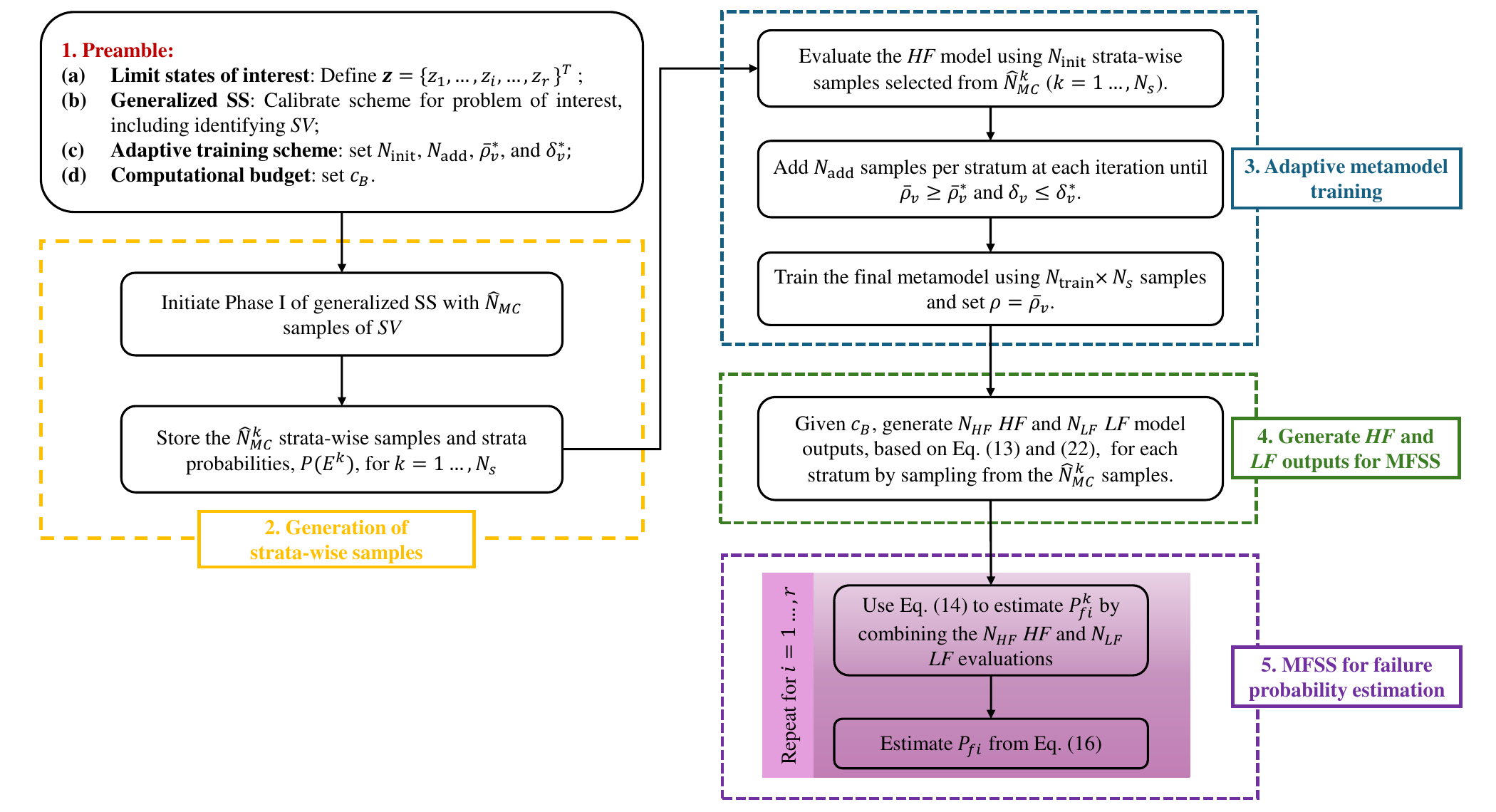}
  \caption{Flowchart illustrating the main parts of the proposed adaptive metamodel-based MFSS scheme.}
  \label{fig: framework}
\end{figure*}

Building on the previous developments, Fig.~\ref{fig: framework} illustrates the overall workflow of the proposed scheme. The key prerequisites include: (a) defining a threshold vector $\boldsymbol{z} = \{z_1,\allowbreak \dots,\allowbreak z_i,\allowbreak \dots,\allowbreak z_r\}^T$ that specifies the limit states of interest; (b) calibrating the GSS scheme of Sec. \ref{subSec: Generalized Stratified Sampling}, including selecting an appropriate $SV$; (c) specifying the variables required for adaptive training, including the number of $HF$ samples for initial training ($N_{\text{init}}$), the number of samples added per iteration ($N_{\text{add}}$), and the target weighted mean correlation and its associated COV, $\bar{\rho}_v^*$ and $\delta_v^*$; and (d) specifying the total computational budget, $c_B$, for MFSS.

The proposed scheme begins by setting up the GSS scheme of Sec.~\ref{subSec: Generalized Stratified Sampling} for the problem of interest. This process results in the selection of a suitable $SV$ and the identification of an appropriate number of strata, $N_s$. With the generalized SS scheme in place, Phase-I sampling is executed with $\hat{N}_{MC}$ MC samples. The adaptive training scheme is then initiated by choosing an equal number of samples, $N_{\text{init}}$, from each stratum to run the $HF$ model. These samples then serve as the initial training dataset for the deep learning metamodel of Sec.~\ref{ssubsec: Deep learning-based metamodel}. If the model correlation does not satisfy the targets $\bar{\rho}_v^*$ and $\delta_v^*$, the next iteration of the training scheme is invoked by adding $N_{\text{add}}$ samples per stratum. This process continues until the model satisfies the targets. The total number of $HF$ training samples used in each stratum is given by $N_{\text{train}} = N_{\text{init}} + N_{\text{add}} \cdot i_{\text{train}}$, where $i_{\text{train}}$ is the number of iterations of the adaptive scheme. The final $LF$ model is developed using all $N_{\text{train}} \times N_s$ $HF$ samples, i.e., across all folds. The correlation coefficient for use within the MFSS setting, $\rho = \bar{\rho}_v$, is that determined at the end of the adaptive training scheme.

Once the $LF$ model is developed, the next step involves generating new $HF$ and $LF$ samples for evaluation of the MFSS estimator. Based on the optimal allocation scheme described in Sec.~\ref{SubSec: MFSS} and a computational budget of $c_B$, the number of stratum-wise $HF$ evaluations, $N_{HF}$, can be determined as:
\begin{equation}
    N_{HF} = \frac{c_B}{r^* \cdot c_{LF} + c_{HF}}
    \label{eq: allocation-p}
\end{equation}
where $r^*$ is defined in Eq.~\eqref{eq: r_critical}, from which the number of $LF$ evaluations per stratum can be determined as $N_{LF} = r^* \cdot N_{HF}$. The strata-wise failure probability is estimated through Eq.~\eqref{eq: mfss-k} by combining $N_{HF}$ $HF$ and $N_{LF}$ $LF$ model evaluations. The overall failure probability across the limit states defined in $\boldsymbol{z}$ is then estimated through Eq.~\eqref{eq: mfss}. To measure the computational efficiency over standard GSS, the speedup, $sp_{MS}$, can be assessed by using Eq.~\eqref{eq: sp_mfss}.


\section{Case Study}
\subsection{High-Fidelity Structural Model and Uncertainties}
\label{subSec: High-fidelity Structural Model and Uncertainties}

\subsubsection{Building System}
\label{subSec:sys}

\begin{figure}[]
  \centering
  \includegraphics[scale=0.7]{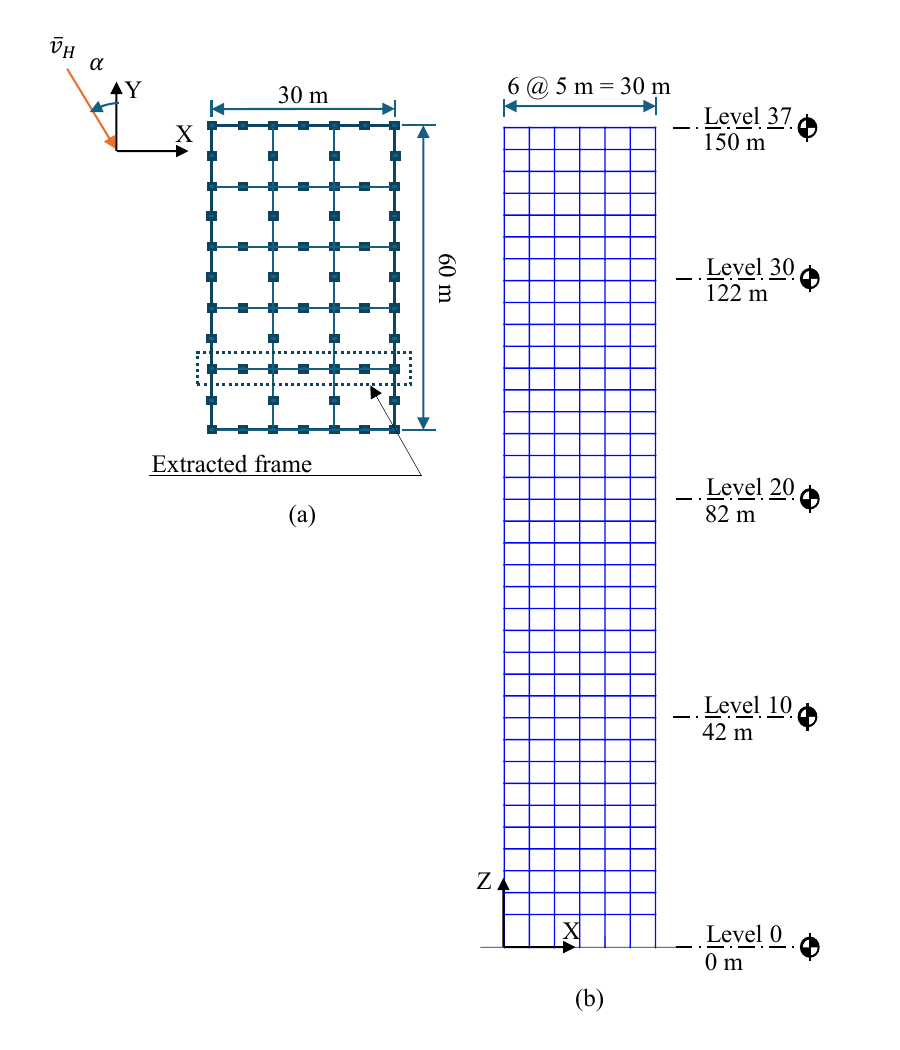}
  \caption{Illustration of the 2D, 37-story steel structural system: (a) plan layout of the building; (b) extracted 2D frame.}
  \label{fig: building}
\end{figure}

To demonstrate the applicability and efficiency of the proposed framework, a case study is conducted on a two-dimensional (2D) 37-story steel moment-resisting frame extracted from a three-dimensional building, as shown in Fig.~\ref{fig: building}. The total height of the structure is 150~m, with a story height of 6~m for the first floor and 4~m for each of the remaining floors. Each floor consists of six spans of equal width (5~m), resulting in a total width of 30~m. The structural system comprises box-section columns and AISC (American Institute of Steel Construction) wide-flange W24 beam sections. All members are composed of steel, with a Young’s modulus of 200 GPa and a yield stress of 355 MPa. The specific members used for the frame are reported in Table~\ref{tab:sections}. In addition to the self-weight of the members, each floor carries an additional mass based on a building density of 100~kg/m\textsuperscript{3}. The archetype system was assumed to be located in a suburban setting in New York City and designed to remain predominantly elastic under a non-directional, site-specific mean hourly wind speed at the building top of 46 m/s, corresponding to a mean recurrence interval (MRI) of approximately 700 years.

The scenario of interest in this work is the extreme alongwind response of the frame when subjected to stochastic wind loads, $\boldsymbol{F}(t; \boldsymbol{\theta})$, as defined in Eq.~(\ref{eq: eom}), over a 10-minute duration. A wind direction of $90^\circ$ was therefore considered, and the stochastic wind loads were calibrated to a 10-minute mean wind speed at the building top, $\bar{v}_H$, of 60 m/s, which corresponded to a MRI of 10,000 years. Strong response nonlinearity is therefore expected. The goal is to characterize the probabilistic response of the system.

\begin{table}[]
\begin{center}
\begin{threeparttable}
\caption{Section sizes used in the steel frame.}
\footnotesize
\setlength{\tabcolsep}{20pt}	
\label{tab:sections}
\begin{tabular}{l c c}
\hline
Floors & Beams & Box columns [cm] \\ \hline
1 - 20 &  W24$\times$192 & 50$\times$2.5 \\
21- 30 &  W24$\times$103 & 40$\times$2.0 \\
31 - 37 &  W24$\times$103 &  35$\times$1.8 \\ \hline
\end{tabular}%
\begin{tablenotes}
\footnotesize
\item Note: Box column size defined as (centerline width) $\times$ (wall thickness).
\end{tablenotes}
\end{threeparttable}
\end{center}
\end{table}


\subsubsection{Stochastic Wind Load Model}
\label{subSec:windmod}

To simulate $\boldsymbol{F}(t; \boldsymbol{\theta})$, the spectral proper orthogonal decomposition model, as outlined in \cite{chuang2019efficient,Duarte2023}, was adopted and calibrated to a dataset corresponding to the building geometry and surrounding conditions of the Tokyo Polytechnic University aerodynamic database \cite{TPU2007winddb}. As described previously, the wind loads were calibrated to a 10-minute mean wind speed of $\bar{v}_H = 60$ m/s at the building top, corresponding to a wind direction of $90^\circ$. Consistent with the extreme loading scenario considered, the total duration of the stochastic wind load realizations was set to 10 minutes. A time step of 0.5 s was adopted, as wind loading can be assumed to have negligible energy content above 1 Hz. To properly simulate the initial and final conditions, the first minute was linearly ramped up, while the final two minutes included a one-minute linear ramp down followed by one minute of zero loading. $\boldsymbol{F}(t; \boldsymbol{\theta})$ was applied laterally in the plane of the frame at each floor level; that is, $\boldsymbol{F}(t; \boldsymbol{\theta})$ is a $37 \times 1$ multivariate stochastic process.
The input uncertainty, $\boldsymbol{\theta}$, consisted of the independent and identically distributed uniform random variables in $[0,2\pi]$, modeling the stochasticity in $\boldsymbol{F}(t; \boldsymbol{\theta})$. 


\subsubsection{High-Fidelity Structural Model}
\label{subSec:HFmod}

For this case study, Eq.~(\ref{eq: eom}) can be written as:
\begin{equation}
\boldsymbol{M}  \ddot{\boldsymbol{y}}(t; \boldsymbol{\theta}) + \boldsymbol{C} \dot{\boldsymbol{y}}(t; \boldsymbol{\theta}) + \boldsymbol{f}_{\text{nl}}(t;\boldsymbol{y}(t; \boldsymbol{\theta}),\dot{\boldsymbol{y}}(t; \boldsymbol{\theta}))  = \boldsymbol{F}(t; \boldsymbol{\theta})
\label{eq:eom-nonlinear}
\end{equation}
where $\boldsymbol{M}$ and $\boldsymbol{C}$ are the mass and damping matrices of the system; $\ddot{\boldsymbol{y}}(t)$, $\dot{\boldsymbol{y}}(t)$, and $\boldsymbol{y}(t)$ denote the stochastic acceleration, velocity, and displacement response trajectories; $\boldsymbol{f}_{\text{nl}}(t)$ represents the nonlinear restoring force; and $\boldsymbol{F}(t; \boldsymbol{\theta})$ is the vector of stochastic wind loads. 

To model $\boldsymbol{f}_{\text{nl}}(t)$, a fiber-discretized nonlinear model was established in OpenSees \cite{mazzoni2006opensees}, which served as the $HF$ model for this application. The model comprised 798 degrees of freedom. All structural components were modeled as displacement-based, fiber-discretized finite elements with five integration points along their length. The Steel02 Giuffré-Menegotto-Pinto model \cite{filippou1983effects} with a strain-hardening ratio of $b_0 = 0.001$ was adopted for each fiber. To model fiber damage due to low-cycle fatigue, the OpenSees fatigue material was wrapped around Steel02, incorporating the linear damage accumulation rule and the modified rainflow cycle algorithm \cite{ballio1995unified}. Large displacement effects were captured using a corotational transformation. Inherent damping was modeled using Rayleigh model, calibrated to provide damping ratios of 2.5\% at the first two natural frequencies, $f_1 = 0.28$~Hz and $f_2 = 0.81$~Hz. 

To solve the responses of the $HF$ model, a Newmark-beta direct integration scheme was adopted. An adaptive nonlinear solver was employed to address potential issues of numerical nonconvergence by considering a succession of algorithms and time steps \cite{li2022metamodeling,li2023reliability}. The procedure begins by attempting a solution using a Newton--Raphson (NR) algorithm with line search and a time step of 0.02 s, with linear interpolation of $\boldsymbol{F}(t)$ to reduce the loading resolution from 0.5 s. If this initial attempt fails to converge, the solver proceeds through the following steps in order: an NR algorithm with line search and a time step of 0.002 s; an NR algorithm with a time step of 0.001 s; and finally, a Broyden algorithm with a time step of 0.001 s. The responses $\boldsymbol{y}(t;\boldsymbol{\theta})$ of the $HF$ model were recorded at a fixed time interval of 0.02 s, which serves as the time resolution of the data used in the following.


\subsection{GRU-Based Adaptive Metamodel}
\label{subSec: GRU-based Adaptive Surrogate Model}

\subsubsection{Training Configuration}
\label{ssubSec: Training Configuration}

To calibrate the $LF$ GRU network-based metamodel of Sec. \ref{ssubsec: Deep learning-based metamodel} to the application of this work, the snapshot matrix, $\boldsymbol{X}$, comprised $n_t = 1,200$ snapshots extracted from the displacement responses of the $HF$ training samples. These snapshots were collected at evenly spaced time intervals. POD modes were extracted by performing SVD on $\boldsymbol{X}$, using a truncation criterion of $\eta=99.999\%$. Subsequently, the full space ($n=798$) was reduced to a three-dimensional space ($n_r=3$) through the transformation matrix $\boldsymbol{\Phi}\in \mathcal{R}^{798\times 3}$, constructed by collecting the first three POD modes. Both the reduced inputs and the reduced outputs were normalized by their average peak value. In applying the wavelet decomposition, the level was set to four to cover 99\% of the energy of the responses \cite{wang2020knowledge}.

The network architecture of the $LF$ metamodel had a GRU layer with 200 hidden units and a dropout layer with probability of 0.5. The network was trained by the widely adopted adaptive moment estimation (Adam) algorithm, with the learning rate set to 0.001. The mean squared error was utilized to evaluate the training performance. To monitor possible overfitting, 10\% of the training set was reserved for monitoring the discrepancy between the training and validation losses. As described in Sec.~\ref{ssubsec: adaptive training}, the approximation quality of the $LF$ model was assessed using 5-fold cross-validation to estimate $\bar{\rho}_v$ and $\delta_v$ of the weighted correlation coefficient defined in Eq.~\eqref{eq: rho}, calibrated to the peak absolute reduced displacement.


\subsubsection{Adaptive Metamodel Training}
\label{ssubSec: Adaptive Metamodel Training}

The GSS scheme for the case study was set up using the elastic resultant base moment, $M_R$, as the $SV$. The elastic dynamic model used to estimate $M_R$ was extracted from the OpenSees model described in Sec.~\ref{subSec:HFmod}. The elastic resultant base moment was chosen as the $SV$, i.e., $SV = M_R$, because it has been shown to be well correlated with the extreme response of dynamically sensitive building systems subjected to extreme winds \cite{xu2024collapse,xu2025multiple}. In addition, the evaluation of $M_R$ is straightforward and extremely computationally efficient---even for high-dimensional systems---as it can be performed using a classical model integration scheme based on digital filters truncated to the first few dynamic modes of the system \cite{spence2013data}. This allows Phase-I sampling to be conducted using large sample sets; in this work, 6,000,000 MC samples were used. These samples were used to identify the distribution of $M_R$ and thereby enable the subsequent partitioning of this distribution into $N_s = 10$ strata. To ensure capture of responses with exceedance probabilities smaller than $10^{-3}$, the lower bound of the final stratum was fixed at an exceedance probability of $10^{-3}$. The lower bound defining the first stratum was taken as zero (i.e., the lower bound of the domain of existence of $M_R$), while the final stratum was considered unbounded from above, ensuring the collectively exhaustive nature of the strata. To enforce mutual exclusivity, the upper bound of each intermediate stratum was set equal to the lower bound of the subsequent stratum. Table~\ref{tab: strata-wise samples} lists the upper and lower bounds, the probability of each stratum, and the number of Phase-I MC samples, $\hat{N}^{k}_{MC}$, falling within each stratum. It can be observed that 5,999 samples fall within the stratum with the smallest probability, ensuring an adequate number of samples for subsequent model evaluations.

\begin{table}[htbp]
\caption{Stratification and corresponding strata probabilities.}
\begin{center}
\footnotesize            
\begin{tabular*}{0.6\textwidth}{@{\extracolsep\fill}lcccc@{}}
\hline
Strata & $M_R^{\text{Lower}}$  & $M_R^{\text{Upper}}$  &$P(E^k)$ & $\hat{N}_{MC}^k$ \\\hline
1 & $0$  & $6.62\times10^5$ & 0.0015 & 9,214\\
2 & $6.62\times10^5$ & $7.29\times10^5$ & 0.0683  & 409,884\\
3 & $7.29\times10^5$ & $7.91\times10^5$ & 0.2880 & 1,727,907\\
4 & $7.91\times10^5$ & $8.48\times10^5$ & 0.3320 & 1,992,292 \\
5 & $8.48\times10^5$ & $9.02\times10^5$ & 0.1916 & 1,149,570\\
6 & $9.02\times10^5$ & $9.53\times10^5$ & 0.0787 & 472,190\\
7 & $9.53\times10^5$ & $1.00\times10^6$ & 0.0275 & 164,928\\
8 & $1.00\times10^6$ & $1.05\times10^6$ & 0.0087 & 52,417\\
9 & $1.05\times10^6$ & $1.09\times10^6$ & 0.0026 & 15,599\\
10 & $1.09\times10^6$ & $\infty$ & 0.0010 & 5,999
\\\hline
\end{tabular*}  
\end{center} 
\label{tab: strata-wise samples}
\end{table}

\begin{figure}[htbp]
  \centering
  \includegraphics[scale=0.65]{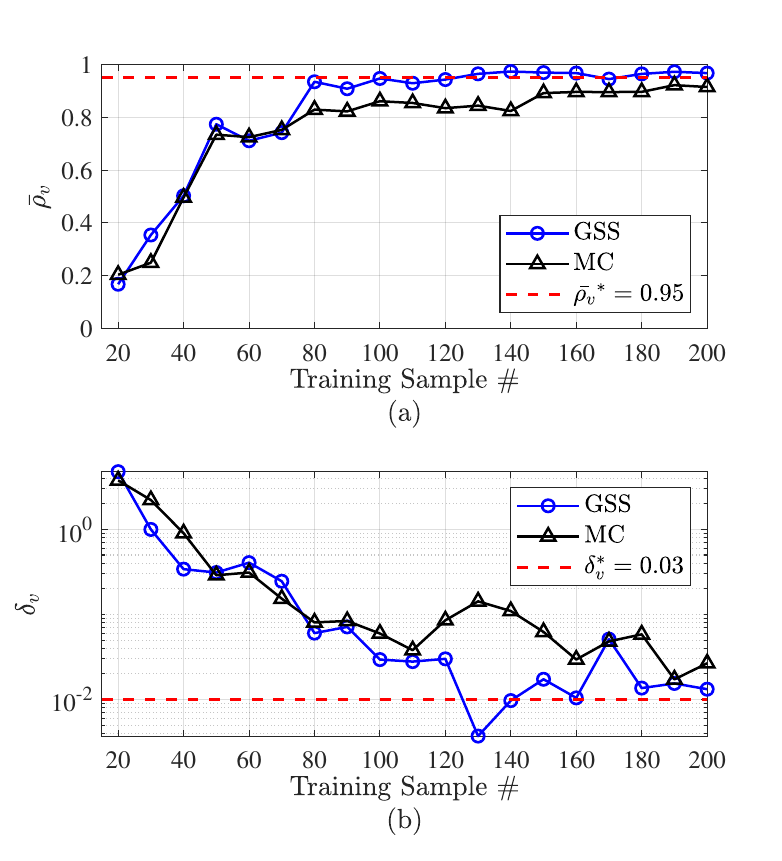}
  \caption{Convergence curves of: (a) the mean of $\rho_v$; and (b) the COV of $\rho_v$, based on training samples selected from sample sets generated using GSS and MC sampling.}
  \label{fig: correlation}
\end{figure}

\begin{figure*}[htbp]
  \centering
  \includegraphics[scale=0.6]{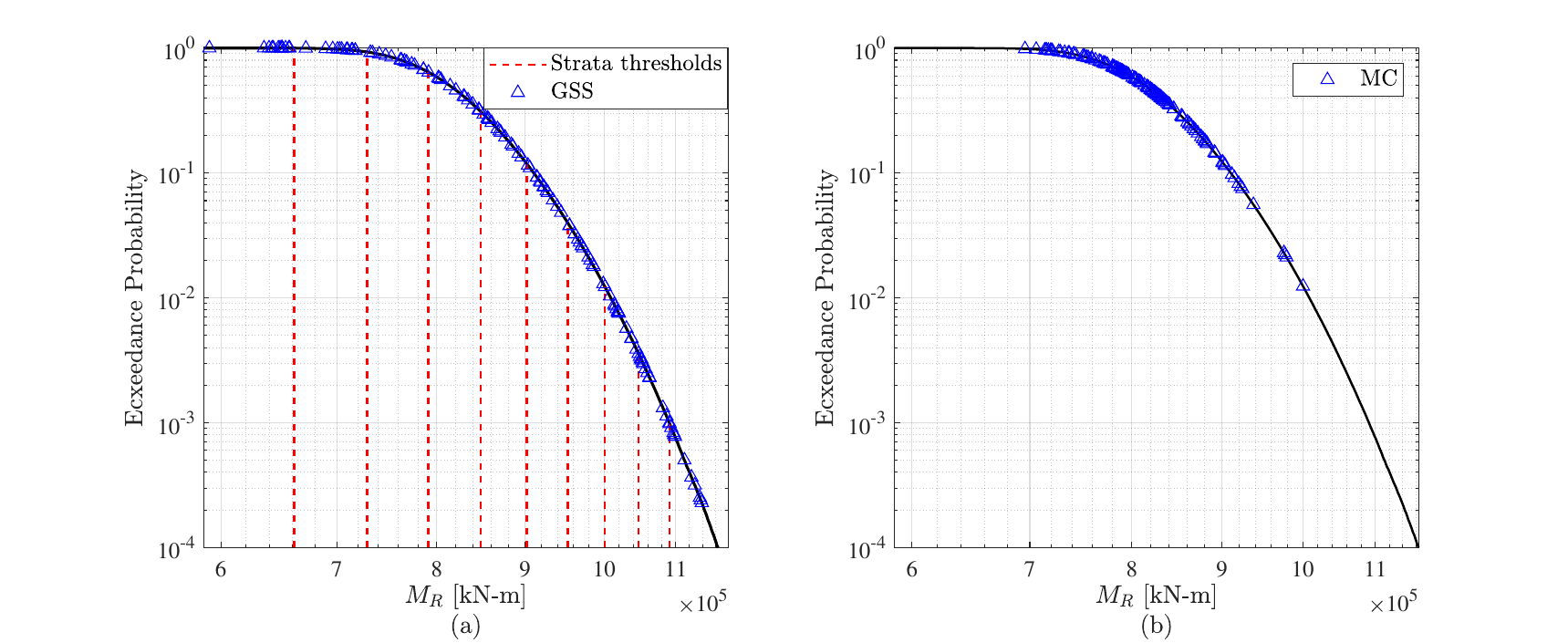}
  \caption{Comparison of sample allocation using: (a) GSS; and (b) MC sampling.}
  \label{fig: allocation}
\end{figure*}

To ensure the approximation quality of the developed $LF$ model, the stopping criteria were set to $\bar{\rho}_v^*=0.95$ and $\delta_v^*=0.03$. The adaptive training scheme was initiated from using $N_{\text{init}}=3$ $HF$ samples in each stratum, resulting in a total of 30 samples. If the criteria was not met, an additional $N_{\text{add}}=1$ random sample from each stratum was added at the next iteration of training until the stopping criteria were satisfied. A total of $N_{\text{train}} \times N_s=130$ samples, as illustrated in Fig. \ref{fig: correlation}, were required to develop the $LF$ model with $\bar{\rho}_v=0.9640$ and $\delta_v=0.37\%$. It can be observed that increasing the number of training samples (e.g., from 130 to 200) does not remarkably enhance model correlation, highlighting the significance of identifying a quasi-optimal number of training samples to balance accuracy and computational efficiency. Fig.~\ref{fig: correlation} also compares the mean and COV of $\rho_v$ for the case in which GSS is used as the basis for selecting training samples, as opposed to simple MC sampling. As can be seen from Fig.~\ref{fig: correlation}, GSS yields faster convergence than MC. This improvement can be attributed to the fact that GSS produces a more diffused sample set, encompassing samples that lead to a wider range of $M_R$ values, and therefore provides more comprehensive information on system responses. This is further illustrated by the sample allocations using the GSS and MC methods, each with 130 samples, as shown in Fig.~\ref{fig: allocation}.

\begin{figure}[t]
  \centering
  \includegraphics[scale=0.7]{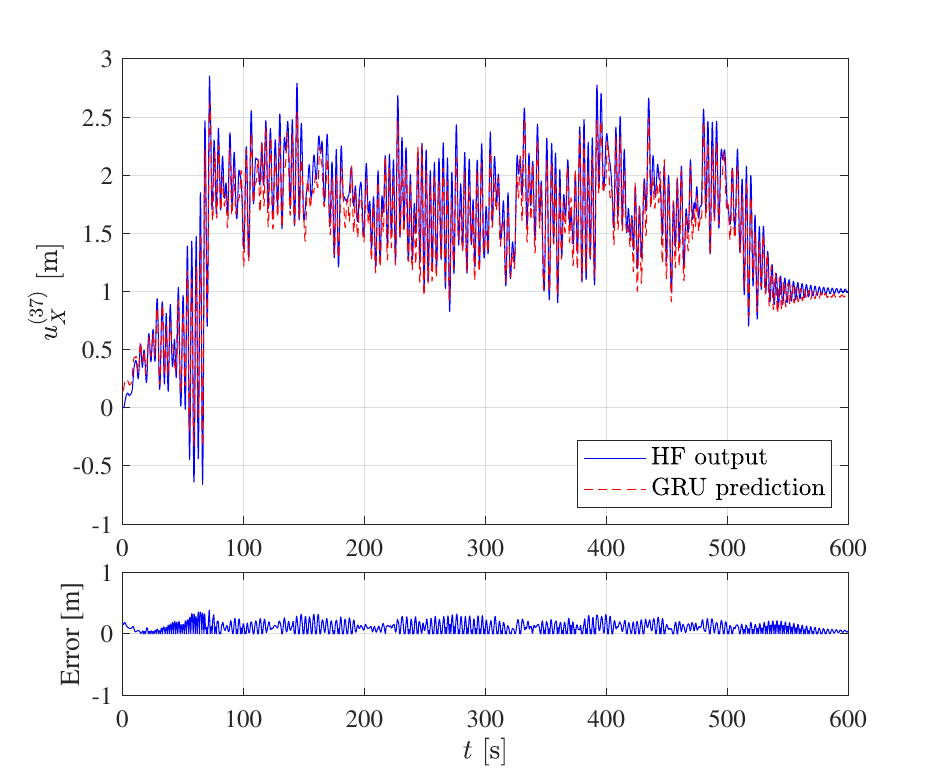}
  \caption{Comparison of the top-floor displacement time history, ${u}_X^{(37)}$, as obtained from the $HF$ model and the GRU-based metamodel for a test sample.}
  \label{fig: disp history}
\end{figure}

To evaluate the performance of the $LF$ model, Fig.~\ref{fig: disp history} compares the time history of the top floor displacement, $u_X^{(37)}$, obtained from the $HF$ model and the GRU-based metamodel for a test sample in the final stratum. While the GRU-based prediction generally captures the time-dependent features of the $HF$ output, it introduces non-negligible errors, reaching up to 10\% in this case. This suggests that directly adopting the data-driven $LF$ model for probabilistic analysis may lead to inaccurate estimations. Increasing the amount of training data can improve the approximation quality of the metamodel, as more information is available during learning. However, this improvement comes with a trade-off: as the training dataset size increases, so does the associated computational cost.

\subsection{Calibration of MFSS and Results}
\label{subSec: Calibration of the Scheme}

To calibrate the proposed scheme, the ratio of computational costs of evaluating the $HF$ and $LF$ models, $c_{HF}/c_{LF}$, was calculated to be 10,000, highlighting the significant computational efficiency of the metamodel compared to the $HF$ model \cite{li2022metamodeling}. The limits states of interest involve peak horizontal displacements at the 10th, 20th, 30th, and 37th floors, denoted as $\hat{u}_X^{(j)}$ where $j$ indicates the number of floors, exceeding thresholds of $\boldsymbol{z} = \{2.5,3.0,4.5,5.0\}^{T}$ m, respectively. To ensure smooth estimation of the failure probability, the consequence measures $h_i(\cdot)$ are assumed to follow a standard normal kernel function.

\begin{figure}[hptb]
  \includegraphics[scale=0.7]{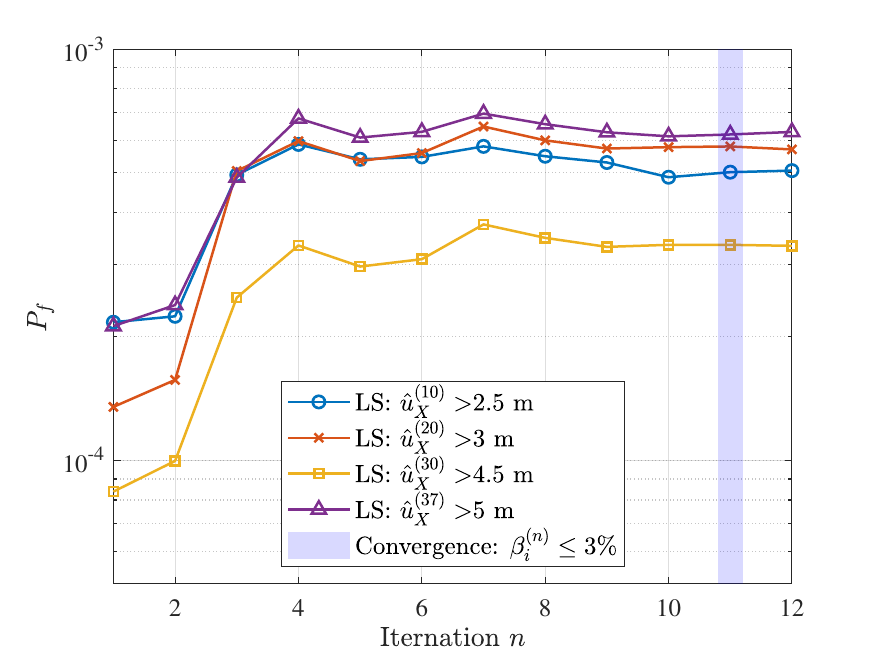}
    \centering
  \caption{Convergence of MFSS estimator when increasing the computational budget}
  \label{fig: convergence of mfss}
\end{figure}

As discussed in Sec.~\ref{subSec:Overall Framework}, one straightforward strategy for allocating $HF$ and $LF$ samples is to predefine the available computational budget, $c_B$. Alternatively, the optimal budget can be identified by monitoring the convergence of the MFSS estimation—a strategy adopted in this case study. In this approach, the number of equally allocated $HF$ samples used in the multi-fidelity estimator of Eq.~(\ref{eq: mfss}) is iteratively increased until a target accuracy is met. For limit state $i$, this accuracy can be evaluated using the following convergence index:
\begin{equation}
\beta_i^{(n)} = \frac{\left| \hat{H}_{i,MS}^{(n+1)} - \hat{H}_{i,MS}^{(n)} \right|}{\hat{H}_{i,MS}^{(n)}}
\label{eq: beta}
\end{equation}
where $n$ is the iteration index, and $\hat{H}_{i,MS}^{(n)}$ denotes the MFSS estimator at the $n$th iteration with a corresponding budget $c_B^{(n)}$. In this application, a single sample was added to each stratum in every iteration. As shown in Fig.~\ref{fig: convergence of mfss}, the MFSS estimation of the probability of failure for each limit state exhibits smooth convergence. In particular, a stopping criterion of $\beta_i^{(n)} \leq 3\%$ was adopted, which was achieved at $N_{HF} = 11$. Following this, the number of $LF$ samples for each stratum was determined to be $N_{LF} = 3,998$, based on the optimal allocation scheme defined by Eq.~\eqref{eq: r_critical}.

\begin{figure}[hptb]
  \centering
  \includegraphics[scale=0.7]{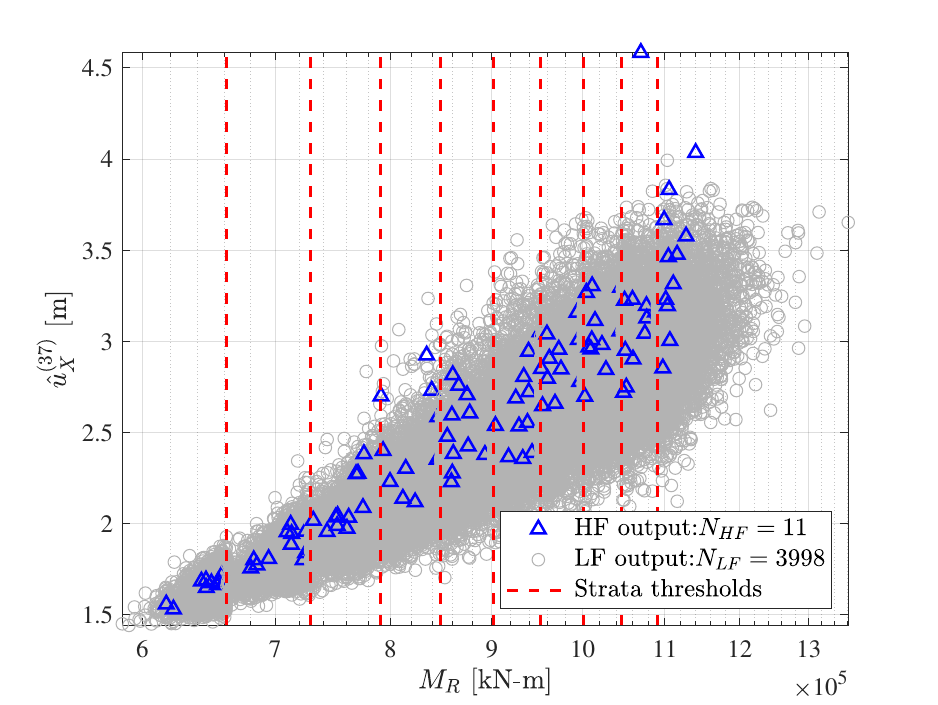}
  \caption{Strata-wise $HF$ and $LF$ evaluations for MFSS}
  \label{fig: mfss all samples}
\end{figure}

Fig.~\ref{fig: mfss all samples} shows the peak top-floor displacements, $\hat{u}_X^{(37)}$, obtained from strata-wise $HF$ and $LF$ samples for the stratification used in this application. It is evident that $\hat{u}_X^{(37)}$ from both the $HF$ and $LF$ models correlate well with $M_R$, verifying the effectiveness of using $M_R$ as the $SV$. To achieve a similar estimator variance using a $HF$-based GSS scheme, $N_{GSS} = 150$, as determined by Eq.~\eqref{eq: N_ss}, samples from each stratum are required. These results are used as a reference to assess the accuracy of the proposed MFSS framework.

\begin{figure*}[]
  \centering
  \includegraphics[scale=0.68]{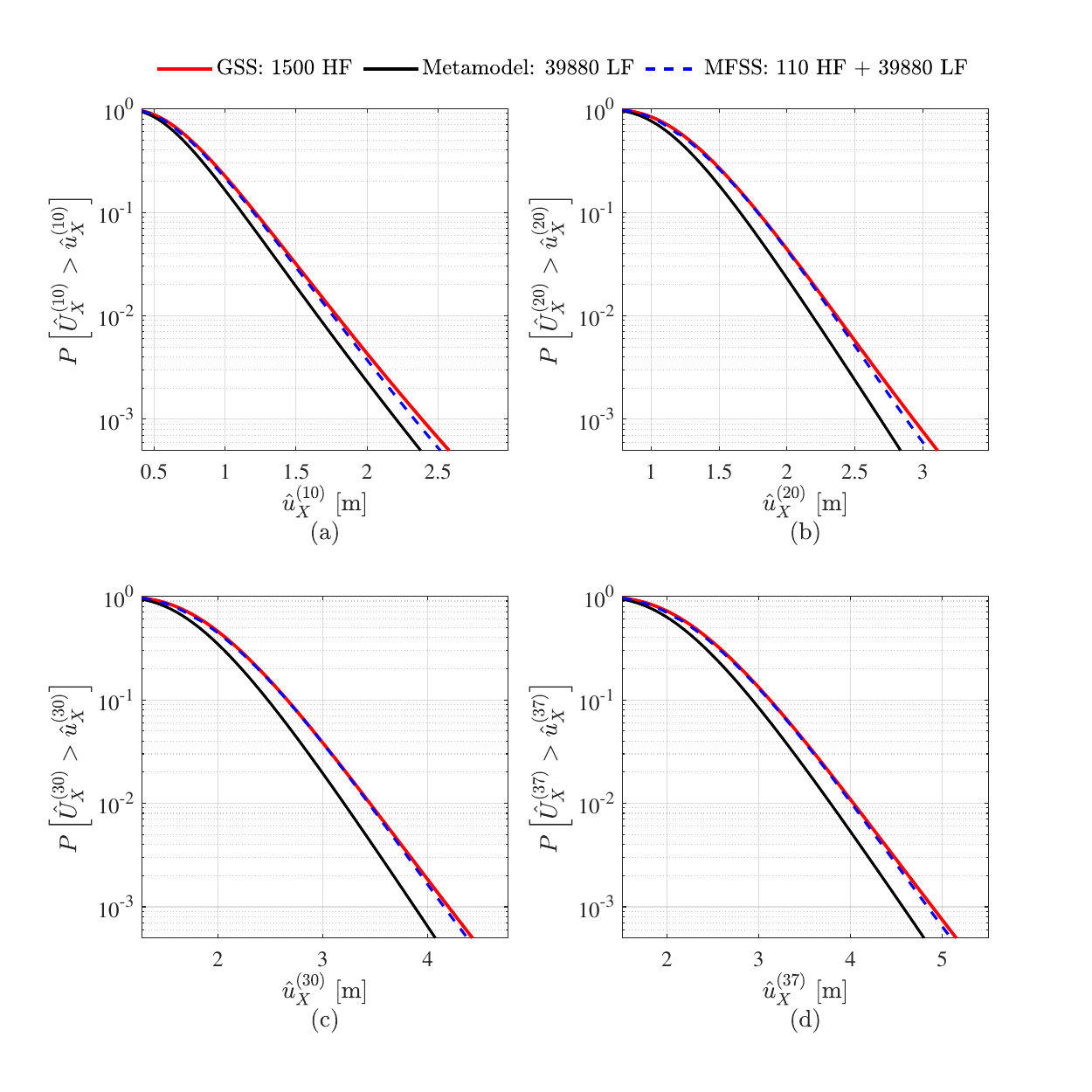}
  \caption{Comparison of peak displacement exceedance probability curves for: (a) the 10\textsuperscript{th} floor; (b) the 20\textsuperscript{th} floor; (c) the 30\textsuperscript{th} floor; and (d) the 37\textsuperscript{th} floor.}
  \label{fig: u exceeding-mfmc}
\end{figure*}

From the above discussion, the MFSS estimator can be established by combining 110 $HF$ and 39,880 $LF$ model evaluations through Eqs.~\eqref{eq: mfss-k} - ~\eqref{eq: mfss}. Table \ref{tab: Pf} compares the estimated failure probabilities and associated COV between HF-based GSS (i.e., 1,500 $HF$ evaluations) and MFSS methods for the limit states of interest. The proposed MFSS scheme shows remarkable accuracy in estimating small failure probabilities, as low as $10^{-4}$, achieving levels of accuracy/variance comparable to the $HF$-based GSS. Additionally, it provides significant computational efficiency with a speed-up of $sp_{MS}$=6.15, using only $16\%$ of the computational budget required for the $HF$-based GSS approach. Fig. \ref{fig: u exceeding-mfmc} shows the exceedance probability curves associated with $\hat{u}_X^{(j)}$, where $j \in \{10, 20, 30, 37\}$, evaluated for different schemes, including GSS using 1,500 $HF$ model evaluations, GSS with 39,880 $LF$ GRU-based outputs, and MFSS. 
\begin{table*}[]
\caption{Comparison of failure probabilities and COV between generalized SS and MFSS for the limit states of interest to this case study.}
\begin{center}
\footnotesize           
\begin{tabular*}{0.9\textwidth}{@{\extracolsep\fill}cccccc@{}}
\hline
\multirow{2}{*}{LS}& \multirow{2}{*}{Description} & \multicolumn{2}{c}{GSS (1500 HF)}  & \multicolumn{2}{c}{MFSS (110 HF+39880 LF)} \\ \cline{3-4} \cline{5-6}
 & & $\hat{H}_{SS}$  & $\kappa_{SS}$  & $\hat{H}_{MS}$  & $\kappa_{MS}$ \\ \hline
{1} & $\hat{u}_X^{(10)}>2.5$ m & $6.65\times 10^{-4}$ & 0.1143 & $5.02\times 10^{-4}$ & 0.0724 \\
{2}& $\hat{u}_X^{(20)}>3.0$ m & $7.62\times 10^{-4}$ & 0.0873 & $5.80\times 10^{-4}$ & 0.0591 \\
{3}& $\hat{u}_X^{(30)}>4.5$ m & $4.01\times 10^{-4}$ & 0.1268 & $3.34\times 10^{-4}$ & 0.0785 \\
{4}& $\hat{u}_X^{(37)}>5.0$ m & $7.48\times 10^{-4}$ & 0.1233 & $6.20\times 10^{-4}$ & 0.1106
\\\hline
\end{tabular*}  
\end{center} 
\label{tab: Pf}
\end{table*}
It can be observed that the MFSS scheme accurately reproduces the exceedance probability curves by integrating a small number of $HF$ model evaluations with a substantial number of $LF$ evaluations. This illustrates the potential of the proposed approach to significantly reduce the computational demand associated with $HF$ model evaluations when assessing small probabilities. Noteworthy, it is evident that exceedance probability curves based solely on GRU-based $LF$ outputs can yield bias, resulting from the fact that the $LF$ model within the MFSS setting generally provides only an approximation of the true response. The MFSS scheme effectively removes this bias by employing a small $HF$ dataset for correction. Overall, the MFSS scheme achieves a balance between accuracy and computational efficiency by leveraging the strengths of both the $HF$ and $LF$ models.


\section{Conclusions}

This paper presented a Multi-Fidelity Stratified Sampling (MFSS) scheme that integrates GSS, MFMC, and adaptive AI-driven metamodeling for efficient estimation of small failure probabilities in high-dimensional, nonlinear structural systems subjected to stochastic excitation. The proposed approach partitions the probability space of a carefully selected stratification variable into multiple strata. A deep learning-based metamodel is trained using $HF$ model evaluations drawn from each stratum, and subsequently used as a computationally efficient $LF$ model within a bi-fidelity framework. To ensure that the $LF$ model maintains sufficient correlation with the $HF$ model, an adaptive training strategy is introduced. This strategy incrementally augments the training dataset until a target correlation threshold with prescribed COV is reached, balancing approximation quality and training cost. Conditional failure probabilities are estimated using MFMC based on an optimal allocation of $HF$ and $LF$ model evaluations across strata. The unconditional failure probability is subsequently computed using the total probability theorem. Application of the MFSS framework to a full-scale high-rise steel building subjected to extreme wind excitation demonstrates the capability of the proposed scheme to estimate exceedance probability curves for multiple limit states involving extreme nonlinear responses, while significantly reducing computational cost compared to GSS based solely on $HF$ model evaluations. By leveraging the strengths of GSS and multi-fidelity modeling, the MFSS scheme provides a scalable framework for efficient estimation of small failure probabilities in complex, nonlinear stochastic systems.


\section*{Acknowledgments}
This research effort was supported in part by the National Science Foundation (NSF) under Grant No. CMMI-2118488. This support is gratefully acknowledged.


\appendix
\section{Unbiasedness of the MFMC Estimator}
\label{appen: MFMC}

The expectation of the MFMC estimator for the probability of failure associated with limit state $i$ can be written as:
\begin{align}   
\mathbb{E}\left[\hat{H}_{i,MF}\right] 
= \mathbb{E}\left[\hat{s}_{HF}^{N_{i,HF}} + a_i\left(\hat{s}_{LF}^{N_{i,LF}} - \hat{s}_{LF}^{N_{i,HF}}\right) \right]
\label{eq: expectation of MFMC estimator}
\end{align}
By applying the linearity of expectation, Eq.~\eqref{eq: expectation of MFMC estimator} can be expressed as:
\begin{align}   
\mathbb{E}\left[\hat{H}_{i,MF}\right] 
= \mathbb{E}\left[\hat{s}_{HF}^{N_{i,HF}}\right] 
+ a_i\left( \mathbb{E}\left[\hat{s}_{LF}^{N_{i,LF}}\right] - \mathbb{E}\left[\hat{s}_{LF}^{N_{i,HF}}\right] \right)
\label{eq: expected MFMC estimator linearity}
\end{align}
Taking advantage of the unbiasedness of the MC estimator, the following holds:
\begin{align}
\mathbb{E}\left[\hat{H}_{i,MF}\right] 
&= \mathbb{E}\left[h_{i,HF}\right] + a_i\left( \mathbb{E}\left[h_{i,LF}\right] - \mathbb{E}\left[h_{i,LF}\right] \right) \nonumber \\
&= \mathbb{E}\left[h_{i,HF}\right]
\label{eq: expected MFMC estimator -unbiased}
\end{align}
where $h_{i,HF}$ and $h_{i,LF}$ represent the consequence measures for limit state $i$ based on the high- and low-fidelity model outputs, respectively. This confirms that the expectation of the MFMC estimator equals the true expectation of the high-fidelity consequence measure, thereby verifying the unbiasedness of the MFMC estimator.

\bibliographystyle{model3-num-names}
\bibliography{Manuscript_R0}

\begin{thebibliography}{70}
\providecommand{\natexlab}[1]{#1}
\providecommand{\url}[1]{\texttt{#1}}
\providecommand{\href}[2]{#2}
\providecommand{\path}[1]{#1}
\providecommand{\eprint}[1]{\href{http://arxiv.org/abs/#1}{\path{#1}}}
\providecommand{\DOIprefix}{doi:}
\providecommand{\ArXivprefix}{arXiv:}
\providecommand{\URLprefix}{URL: }
\providecommand{\Pubmedprefix}{pmid:}
\providecommand{\doi}[1]{\href{http://dx.doi.org/#1}{\path{#1}}}
\providecommand{\Pubmed}[1]{\href{pmid:#1}{\path{#1}}}
\providecommand{\BIBand}{and}
\providecommand{\bibinfo}[2]{#2}
\ifx\xfnm\undefined \def\xfnm[#1]{\unskip,\space#1}\fi
\bibitem[{Koutsourelakis et~al.(2004)Koutsourelakis, Pradlwarter and Schueller}]{koutsourelakis2004reliability}
\bibinfo{author}{Koutsourelakis\xfnm[ P.S.]}, \bibinfo{author}{Pradlwarter\xfnm[ H.J.]}, \bibinfo{author}{Schueller\xfnm[ G.I.]}.
\newblock \bibinfo{title}{Reliability of structures in high dimensions, part {I}: Algorithms and applications}.
\newblock \bibinfo{journal}{Probabilistic Engineering Mechanics} \bibinfo{year}{2004};\bibinfo{volume}{19}(\bibinfo{number}{4}):\bibinfo{pages}{409--417}.
\bibitem[{Beck et~al.(2014)Beck, Kougioumtzoglou and Dos~Santos}]{beck2014optimal}
\bibinfo{author}{Beck\xfnm[ A.T.]}, \bibinfo{author}{Kougioumtzoglou\xfnm[ I.A.]}, \bibinfo{author}{Dos~Santos\xfnm[ K.R.M.]}.
\newblock \bibinfo{title}{Optimal performance-based design of non-linear stochastic dynamical {RC} structures subject to stationary wind excitation}.
\newblock \bibinfo{journal}{Engineering Structures} \bibinfo{year}{2014};\bibinfo{volume}{78}:\bibinfo{pages}{145--153}.
\bibitem[{Shields and Sundar(2015)}]{shields2015targeted}
\bibinfo{author}{Shields\xfnm[ M.D.]}, \bibinfo{author}{Sundar\xfnm[ V.S.]}.
\newblock \bibinfo{title}{Targeted random sampling: A new approach for efficient reliability estimation for complex systems}.
\newblock \bibinfo{journal}{International Journal of Reliability and Safety} \bibinfo{year}{2015};\bibinfo{volume}{9}(\bibinfo{number}{2--3}):\bibinfo{pages}{174--190}.
\bibitem[{Melchers and Beck(2018)}]{melchers2018structural}
\bibinfo{author}{Melchers\xfnm[ R.E.]}, \bibinfo{author}{Beck\xfnm[ A.T.]}.
\newblock \bibinfo{title}{Structural Reliability Analysis and Prediction}.
\newblock \bibinfo{edition}{3} ed.; \bibinfo{publisher}{John Wiley \& Sons Ltd.}; \bibinfo{year}{2018}.
\bibitem[{Yi et~al.(2018)Yi, Wang and Song}]{yi2018xgaussian}
\bibinfo{author}{Yi\xfnm[ S.r.]}, \bibinfo{author}{Wang\xfnm[ Z.]}, \bibinfo{author}{Song\xfnm[ J.]}.
\newblock \bibinfo{title}{Bivariate {G}aussian mixture–based equivalent linearization method for stochastic seismic analysis of nonlinear structures}.
\newblock \bibinfo{journal}{Earthquake Engineering \& Structural Dynamics} \bibinfo{year}{2018};\bibinfo{volume}{47}:\bibinfo{pages}{678--696}.
\bibitem[{Arunachalam and Spence(2022)}]{arunachalam2022reliability}
\bibinfo{author}{Arunachalam\xfnm[ S.]}, \bibinfo{author}{Spence\xfnm[ S.M.J.]}.
\newblock \bibinfo{title}{Reliability-based collapse assessment of wind-excited steel structures within performance-based wind engineering}.
\newblock \bibinfo{journal}{Journal of Structural Engineering} \bibinfo{year}{2022};\bibinfo{volume}{148}(\bibinfo{number}{9}):\bibinfo{pages}{04022132}.
\bibitem[{Chuang and Spence(2022)}]{chuang2022framework}
\bibinfo{author}{Chuang\xfnm[ W.C.]}, \bibinfo{author}{Spence\xfnm[ S.M.J.]}.
\newblock \bibinfo{title}{A framework for the efficient reliability assessment of inelastic wind-excited structures at dynamic shakedown}.
\newblock \bibinfo{journal}{Journal of Wind Engineering and Industrial Aerodynamics} \bibinfo{year}{2022};\bibinfo{volume}{220}:\bibinfo{pages}{104834}.
\bibitem[{Beck et~al.(2022)Beck, Bosse and Rodrigues}]{beck2022structural}
\bibinfo{author}{Beck\xfnm[ A.T.]}, \bibinfo{author}{Bosse\xfnm[ R.M.]}, \bibinfo{author}{Rodrigues\xfnm[ I.D.]}.
\newblock \bibinfo{title}{On the ergodicity assumption in performance-based engineering}.
\newblock \bibinfo{journal}{Structural Safety} \bibinfo{year}{2022};\bibinfo{volume}{97}:\bibinfo{pages}{102218}.
\bibitem[{Arunachalam and Spence(2023{\natexlab{a}})}]{arunachalam2023generalized}
\bibinfo{author}{Arunachalam\xfnm[ S.]}, \bibinfo{author}{Spence\xfnm[ S.M.J.]}.
\newblock \bibinfo{title}{Generalized stratified sampling for efficient reliability assessment of structures against natural hazards}.
\newblock \bibinfo{journal}{Journal of Engineering Mechanics} \bibinfo{year}{2023}{\natexlab{a}};\bibinfo{volume}{149}(\bibinfo{number}{7}):\bibinfo{pages}{04023042}.
\bibitem[{Goswami et~al.(2025)Goswami, Giovanis, Li, Spence and Shields}]{goswami2025neural}
\bibinfo{author}{Goswami\xfnm[ S.]}, \bibinfo{author}{Giovanis\xfnm[ D.G.]}, \bibinfo{author}{Li\xfnm[ B.]}, \bibinfo{author}{Spence\xfnm[ S.M.J.]}, \bibinfo{author}{Shields\xfnm[ M.D.]}.
\newblock \bibinfo{title}{Neural operators for stochastic modeling of nonlinear structural system response to natural hazards}.
\newblock \bibinfo{howpublished}{arXiv preprint}; \bibinfo{year}{2025}.
\newblock \URLprefix \url{https://arxiv.org/abs/2502.11279}; \bibinfo{note}{arXiv:2502.11279}.
\bibitem[{Deodatis and Shields(2025)}]{deodatis2025spectral}
\bibinfo{author}{Deodatis\xfnm[ G.]}, \bibinfo{author}{Shields\xfnm[ M.D.]}.
\newblock \bibinfo{title}{The spectral representation method: A framework for simulation of stochastic processes, fields, and waves}.
\newblock \bibinfo{journal}{Reliability Engineering \& System Safety} \bibinfo{year}{2025};\bibinfo{volume}{254}:\bibinfo{pages}{110522}.
\bibitem[{Lee et~al.(2025)Lee, Wang and Song}]{lee2025efficient}
\bibinfo{author}{Lee\xfnm[ D.]}, \bibinfo{author}{Wang\xfnm[ Z.]}, \bibinfo{author}{Song\xfnm[ J.]}.
\newblock \bibinfo{title}{Efficient seismic reliability and fragility analysis of lifeline networks using subset simulation}.
\newblock \bibinfo{journal}{Reliability Engineering \& System Safety} \bibinfo{year}{2025};\bibinfo{volume}{260}:\bibinfo{pages}{110947}.
\bibitem[{Giovanis et~al.(2025)Giovanis, Taflanidis and Shields}]{giovanis2025accelerating}
\bibinfo{author}{Giovanis\xfnm[ D.G.]}, \bibinfo{author}{Taflanidis\xfnm[ A.]}, \bibinfo{author}{Shields\xfnm[ M.D.]}.
\newblock \bibinfo{title}{Accelerating uncertainty quantification in incremental dynamic analysis using dimension reduction-based surrogate modeling}.
\newblock \bibinfo{journal}{Bulletin of Earthquake Engineering} \bibinfo{year}{2025};\bibinfo{volume}{23}(\bibinfo{number}{1}):\bibinfo{pages}{391--410}.
\bibitem[{Melchers(1989)}]{melchers1989importance}
\bibinfo{author}{Melchers\xfnm[ R.E.]}.
\newblock \bibinfo{title}{Importance sampling in structural systems}.
\newblock \bibinfo{journal}{Structural Safety} \bibinfo{year}{1989};\bibinfo{volume}{6}(\bibinfo{number}{1}):\bibinfo{pages}{3--10}.
\bibitem[{Au and Beck(2003)}]{au2003important}
\bibinfo{author}{Au\xfnm[ S.K.]}, \bibinfo{author}{Beck\xfnm[ J.L.]}.
\newblock \bibinfo{title}{Important sampling in high dimensions}.
\newblock \bibinfo{journal}{Structural Safety} \bibinfo{year}{2003};\bibinfo{volume}{25}(\bibinfo{number}{2}):\bibinfo{pages}{139--163}.
\bibitem[{Arunachalam and Spence(2023{\natexlab{b}})}]{arunachalam2023efficient}
\bibinfo{author}{Arunachalam\xfnm[ S.]}, \bibinfo{author}{Spence\xfnm[ S.M.J.]}.
\newblock \bibinfo{title}{An efficient stratified sampling scheme for the simultaneous estimation of small failure probabilities in wind engineering applications}.
\newblock \bibinfo{journal}{Structural Safety} \bibinfo{year}{2023}{\natexlab{b}};\bibinfo{volume}{101}:\bibinfo{pages}{102310}.
\bibitem[{Xu and Spence(2024)}]{xu2024collapse}
\bibinfo{author}{Xu\xfnm[ L.]}, \bibinfo{author}{Spence\xfnm[ S.M.J.]}.
\newblock \bibinfo{title}{Collapse reliability of wind-excited reinforced concrete structures by stratified sampling and nonlinear dynamic analysis}.
\newblock \bibinfo{journal}{Reliability Engineering \& System Safety} \bibinfo{year}{2024};:\bibinfo{pages}{110244}.
\bibitem[{Au and Beck(2001)}]{au2001estimation}
\bibinfo{author}{Au\xfnm[ S.K.]}, \bibinfo{author}{Beck\xfnm[ J.L.]}.
\newblock \bibinfo{title}{Estimation of small failure probabilities in high dimensions by subset simulation}.
\newblock \bibinfo{journal}{Probabilistic Engineering Mechanics} \bibinfo{year}{2001};\bibinfo{volume}{16}(\bibinfo{number}{4}):\bibinfo{pages}{263--277}.
\bibitem[{Lucia et~al.(2004)Lucia, Beran and Silva}]{lucia2004reduced}
\bibinfo{author}{Lucia\xfnm[ D.J.]}, \bibinfo{author}{Beran\xfnm[ P.S.]}, \bibinfo{author}{Silva\xfnm[ W.A.]}.
\newblock \bibinfo{title}{Reduced-order modeling: New approaches for computational physics}.
\newblock \bibinfo{journal}{Progress in Aerospace Sciences} \bibinfo{year}{2004};\bibinfo{volume}{40}(\bibinfo{number}{1--2}):\bibinfo{pages}{51--117}.
\bibitem[{Patsialis and Taflanidis(2020)}]{patsialis2020reduced}
\bibinfo{author}{Patsialis\xfnm[ D.]}, \bibinfo{author}{Taflanidis\xfnm[ A.A.]}.
\newblock \bibinfo{title}{Reduced order modeling of hysteretic structural response and applications to seismic risk assessment}.
\newblock \bibinfo{journal}{Engineering Structures} \bibinfo{year}{2020};\bibinfo{volume}{209}:\bibinfo{pages}{110135}.
\bibitem[{Li and Xiu(2010)}]{li2010evaluation}
\bibinfo{author}{Li\xfnm[ J.]}, \bibinfo{author}{Xiu\xfnm[ D.]}.
\newblock \bibinfo{title}{Evaluation of failure probability via surrogate models}.
\newblock \bibinfo{journal}{Journal of Computational Physics} \bibinfo{year}{2010};\bibinfo{volume}{229}(\bibinfo{number}{23}):\bibinfo{pages}{8966--8980}.
\bibitem[{Gidaris et~al.(2015)Gidaris, Taflanidis and Mavroeidis}]{gidaris2015kriging}
\bibinfo{author}{Gidaris\xfnm[ I.]}, \bibinfo{author}{Taflanidis\xfnm[ A.A.]}, \bibinfo{author}{Mavroeidis\xfnm[ G.P.]}.
\newblock \bibinfo{title}{Kriging metamodeling in seismic risk assessment based on stochastic ground motion models}.
\newblock \bibinfo{journal}{Earthquake Engineering \& Structural Dynamics} \bibinfo{year}{2015};\bibinfo{volume}{44}(\bibinfo{number}{14}):\bibinfo{pages}{2377--2399}.
\bibitem[{Li et~al.(2020)Li, Wang and Jia}]{li2020efficient}
\bibinfo{author}{Li\xfnm[ M.]}, \bibinfo{author}{Wang\xfnm[ R.Q.]}, \bibinfo{author}{Jia\xfnm[ G.]}.
\newblock \bibinfo{title}{Efficient dimension reduction and surrogate-based sensitivity analysis for expensive models with high-dimensional outputs}.
\newblock \bibinfo{journal}{Reliability Engineering \& System Safety} \bibinfo{year}{2020};\bibinfo{volume}{195}:\bibinfo{pages}{106725}.
\bibitem[{Lagaros and Papadrakakis(2012)}]{lagaros2012neural}
\bibinfo{author}{Lagaros\xfnm[ N.D.]}, \bibinfo{author}{Papadrakakis\xfnm[ M.]}.
\newblock \bibinfo{title}{Neural network-based prediction schemes of the nonlinear seismic response of {3D} buildings}.
\newblock \bibinfo{journal}{Advances in Engineering Software} \bibinfo{year}{2012};\bibinfo{volume}{44}(\bibinfo{number}{1}):\bibinfo{pages}{92--115}.
\bibitem[{Sharma et~al.(2024)Sharma, Nov{\'a}k and Shields}]{sharma2024physics}
\bibinfo{author}{Sharma\xfnm[ H.]}, \bibinfo{author}{Nov{\'a}k\xfnm[ L.]}, \bibinfo{author}{Shields\xfnm[ M.]}.
\newblock \bibinfo{title}{Physics-constrained polynomial chaos expansion for scientific machine learning and uncertainty quantification}.
\newblock \bibinfo{journal}{Computer Methods in Applied Mechanics and Engineering} \bibinfo{year}{2024};\bibinfo{volume}{431}:\bibinfo{pages}{117314}.
\bibitem[{Li et~al.(2024)Li, Arunachalam and Spence}]{li2024multi}
\bibinfo{author}{Li\xfnm[ M.]}, \bibinfo{author}{Arunachalam\xfnm[ S.]}, \bibinfo{author}{Spence\xfnm[ S.M.J.]}.
\newblock \bibinfo{title}{A multi-fidelity stochastic simulation scheme for estimation of small failure probabilities}.
\newblock \bibinfo{journal}{Structural Safety} \bibinfo{year}{2024};\bibinfo{volume}{106}:\bibinfo{pages}{102397}.
\bibitem[{Ng and Willcox(2014)}]{ng2014multifidelity}
\bibinfo{author}{Ng\xfnm[ L.W.T.]}, \bibinfo{author}{Willcox\xfnm[ K.E.]}.
\newblock \bibinfo{title}{Multifidelity approaches for optimization under uncertainty}.
\newblock \bibinfo{journal}{International Journal for Numerical Methods in Engineering} \bibinfo{year}{2014};\bibinfo{volume}{100}(\bibinfo{number}{10}):\bibinfo{pages}{746--772}.
\bibitem[{Peherstorfer et~al.(2016{\natexlab{a}})Peherstorfer, Cui, Marzouk and Willcox}]{peherstorfer2016multifidelity}
\bibinfo{author}{Peherstorfer\xfnm[ B.]}, \bibinfo{author}{Cui\xfnm[ T.]}, \bibinfo{author}{Marzouk\xfnm[ Y.]}, \bibinfo{author}{Willcox\xfnm[ K.]}.
\newblock \bibinfo{title}{Multifidelity importance sampling}.
\newblock \bibinfo{journal}{Computer Methods in Applied Mechanics and Engineering} \bibinfo{year}{2016}{\natexlab{a}};\bibinfo{volume}{300}:\bibinfo{pages}{490--509}.
\bibitem[{Peherstorfer et~al.(2018)Peherstorfer, Willcox and Gunzburger}]{peherstorfer2018survey}
\bibinfo{author}{Peherstorfer\xfnm[ B.]}, \bibinfo{author}{Willcox\xfnm[ K.]}, \bibinfo{author}{Gunzburger\xfnm[ M.]}.
\newblock \bibinfo{title}{Survey of multifidelity methods in uncertainty propagation, inference, and optimization}.
\newblock \bibinfo{journal}{SIAM Review} \bibinfo{year}{2018};\bibinfo{volume}{60}(\bibinfo{number}{3}):\bibinfo{pages}{550--591}.
\bibitem[{Nelson(1987)}]{nelson1987control}
\bibinfo{author}{Nelson\xfnm[ B.L.]}.
\newblock \bibinfo{title}{On control variate estimators}.
\newblock \bibinfo{journal}{Computers \& Operations Research} \bibinfo{year}{1987};\bibinfo{volume}{14}(\bibinfo{number}{3}):\bibinfo{pages}{219--225}.
\bibitem[{Han et~al.(2024)Han, Kramer, Lee, Narayan and Xu}]{han2024approximate}
\bibinfo{author}{Han\xfnm[ R.]}, \bibinfo{author}{Kramer\xfnm[ B.]}, \bibinfo{author}{Lee\xfnm[ D.]}, \bibinfo{author}{Narayan\xfnm[ A.]}, \bibinfo{author}{Xu\xfnm[ Y.]}.
\newblock \bibinfo{title}{An approximate control variates approach to multifidelity distribution estimation}.
\newblock \bibinfo{journal}{SIAM/ASA Journal on Uncertainty Quantification} \bibinfo{year}{2024};\bibinfo{volume}{12}(\bibinfo{number}{4}):\bibinfo{pages}{1349--1388}.
\bibitem[{Cliffe et~al.(2011)Cliffe, Giles, Scheichl and Teckentrup}]{cliffe2011multilevel}
\bibinfo{author}{Cliffe\xfnm[ K.A.]}, \bibinfo{author}{Giles\xfnm[ M.B.]}, \bibinfo{author}{Scheichl\xfnm[ R.]}, \bibinfo{author}{Teckentrup\xfnm[ A.L.]}.
\newblock \bibinfo{title}{Multilevel {M}onte {C}arlo methods and applications to elliptic {PDEs} with random coefficients}.
\newblock \bibinfo{journal}{Computing and Visualization in Science} \bibinfo{year}{2011};\bibinfo{volume}{14}:\bibinfo{pages}{3--15}.
\bibitem[{Giles(2015)}]{giles2015multilevel}
\bibinfo{author}{Giles\xfnm[ M.B.]}.
\newblock \bibinfo{title}{Multilevel {M}onte {C}arlo methods}.
\newblock \bibinfo{journal}{Acta Numerica} \bibinfo{year}{2015};\bibinfo{volume}{24}:\bibinfo{pages}{259--328}.
\bibitem[{Peherstorfer et~al.(2016{\natexlab{b}})Peherstorfer, Willcox and Gunzburger}]{peherstorfer2016optimal}
\bibinfo{author}{Peherstorfer\xfnm[ B.]}, \bibinfo{author}{Willcox\xfnm[ K.]}, \bibinfo{author}{Gunzburger\xfnm[ M.]}.
\newblock \bibinfo{title}{Optimal model management for multifidelity {M}onte {C}arlo estimation}.
\newblock \bibinfo{journal}{SIAM Journal on Scientific Computing} \bibinfo{year}{2016}{\natexlab{b}};\bibinfo{volume}{38}(\bibinfo{number}{5}):\bibinfo{pages}{A3163--A3194}.
\bibitem[{Kramer et~al.(2019)Kramer, Marques, Peherstorfer, Villa and Willcox}]{kramer2019multifidelity}
\bibinfo{author}{Kramer\xfnm[ B.]}, \bibinfo{author}{Marques\xfnm[ A.N.]}, \bibinfo{author}{Peherstorfer\xfnm[ B.]}, \bibinfo{author}{Villa\xfnm[ U.]}, \bibinfo{author}{Willcox\xfnm[ K.]}.
\newblock \bibinfo{title}{Multifidelity probability estimation via fusion of estimators}.
\newblock \bibinfo{journal}{Journal of Computational Physics} \bibinfo{year}{2019};\bibinfo{volume}{392}:\bibinfo{pages}{385--402}.
\bibitem[{Yi et~al.(2021)Yi, Wu, Zhou, Cheng, Ling and Liu}]{yi2021active}
\bibinfo{author}{Yi\xfnm[ J.]}, \bibinfo{author}{Wu\xfnm[ F.]}, \bibinfo{author}{Zhou\xfnm[ Q.]}, \bibinfo{author}{Cheng\xfnm[ Y.]}, \bibinfo{author}{Ling\xfnm[ H.]}, \bibinfo{author}{Liu\xfnm[ J.]}.
\newblock \bibinfo{title}{An active-learning method based on multi-fidelity kriging model for structural reliability analysis}.
\newblock \bibinfo{journal}{Structural and Multidisciplinary Optimization} \bibinfo{year}{2021};\bibinfo{volume}{63}:\bibinfo{pages}{173--195}.
\bibitem[{Renganathan et~al.(2022)Renganathan, Rao and Navon}]{renganathan2022multifidelity}
\bibinfo{author}{Renganathan\xfnm[ A.]}, \bibinfo{author}{Rao\xfnm[ V.]}, \bibinfo{author}{Navon\xfnm[ I.]}.
\newblock \bibinfo{title}{Multifidelity {G}aussian processes for failure boundary and probability estimation}.
\newblock In: \bibinfo{booktitle}{AIAA SCITECH 2022 Forum}. \bibinfo{year}{2022}, p. \bibinfo{pages}{0390}.
\bibitem[{Patsialis and Taflanidis(2021)}]{patsialis2021multi}
\bibinfo{author}{Patsialis\xfnm[ D.]}, \bibinfo{author}{Taflanidis\xfnm[ A.A.]}.
\newblock \bibinfo{title}{Multi-fidelity {M}onte {C}arlo for seismic risk assessment applications}.
\newblock \bibinfo{journal}{Structural Safety} \bibinfo{year}{2021};\bibinfo{volume}{93}:\bibinfo{pages}{102129}.
\bibitem[{Jung et~al.(2024)Jung, Taflanidis, Kyprioti and Zhang}]{jung2024adaptive}
\bibinfo{author}{Jung\xfnm[ W.]}, \bibinfo{author}{Taflanidis\xfnm[ A.A.]}, \bibinfo{author}{Kyprioti\xfnm[ A.P.]}, \bibinfo{author}{Zhang\xfnm[ J.]}.
\newblock \bibinfo{title}{Adaptive multi-fidelity {M}onte {C}arlo for real-time probabilistic storm surge predictions}.
\newblock \bibinfo{journal}{Reliability Engineering \& System Safety} \bibinfo{year}{2024};\bibinfo{volume}{247}:\bibinfo{pages}{109994}.
\bibitem[{Simonoff(2012)}]{simonoff2012smoothing}
\bibinfo{author}{Simonoff\xfnm[ J.S.]}.
\newblock \bibinfo{title}{Smoothing Methods in Statistics}.
\newblock \bibinfo{publisher}{Springer Science \& Business Media}; \bibinfo{year}{2012}.
\bibitem[{Suksuwan and Spence(2018)}]{suksuwan2018optimization}
\bibinfo{author}{Suksuwan\xfnm[ A.]}, \bibinfo{author}{Spence\xfnm[ S.M.J.]}.
\newblock \bibinfo{title}{Optimization of uncertain structures subject to stochastic wind loads under system-level first excursion constraints: A data-driven approach}.
\newblock \bibinfo{journal}{Computers \& Structures} \bibinfo{year}{2018};\bibinfo{volume}{210}:\bibinfo{pages}{58--68}.
\bibitem[{Silverman(2018)}]{silverman2018density}
\bibinfo{author}{Silverman\xfnm[ B.W.]}.
\newblock \bibinfo{title}{Density Estimation for Statistics and Data Analysis}.
\newblock \bibinfo{publisher}{Routledge}; \bibinfo{year}{2018}.
\bibitem[{Leli{\`e}vre et~al.(2018)Leli{\`e}vre, Beaurepaire, Mattrand and Gayton}]{lelievre2018ak}
\bibinfo{author}{Leli{\`e}vre\xfnm[ N.]}, \bibinfo{author}{Beaurepaire\xfnm[ P.]}, \bibinfo{author}{Mattrand\xfnm[ C.]}, \bibinfo{author}{Gayton\xfnm[ N.]}.
\newblock \bibinfo{title}{{AK-MCSi}: A {Kriging}-based method to deal with small failure probabilities and time-consuming models}.
\newblock \bibinfo{journal}{Structural Safety} \bibinfo{year}{2018};\bibinfo{volume}{73}:\bibinfo{pages}{1--11}.
\bibitem[{Li and Spence(2022)}]{li2022metamodeling}
\bibinfo{author}{Li\xfnm[ B.]}, \bibinfo{author}{Spence\xfnm[ S.M.J.]}.
\newblock \bibinfo{title}{Metamodeling through deep learning of high-dimensional dynamic nonlinear systems driven by general stochastic excitation}.
\newblock \bibinfo{journal}{Journal of Structural Engineering} \bibinfo{year}{2022};\bibinfo{volume}{148}(\bibinfo{number}{11}):\bibinfo{pages}{04022186}.
\bibitem[{Bamer et~al.(2017)Bamer, Kazemi~Amiri and Bucher}]{bamer2017new}
\bibinfo{author}{Bamer\xfnm[ F.]}, \bibinfo{author}{Kazemi~Amiri\xfnm[ A.]}, \bibinfo{author}{Bucher\xfnm[ C.]}.
\newblock \bibinfo{title}{A new model order reduction strategy adapted to nonlinear problems in earthquake engineering}.
\newblock \bibinfo{journal}{Earthquake Engineering \& Structural Dynamics} \bibinfo{year}{2017};\bibinfo{volume}{46}(\bibinfo{number}{4}):\bibinfo{pages}{537--559}.
\bibitem[{Li et~al.(2021)Li, Chuang and Spence}]{Li2021Response}
\bibinfo{author}{Li\xfnm[ B.]}, \bibinfo{author}{Chuang\xfnm[ W.C.]}, \bibinfo{author}{Spence\xfnm[ S.M.J.]}.
\newblock \bibinfo{title}{Response estimation of multi-degree-of-freedom nonlinear stochastic structural systems through metamodeling}.
\newblock \bibinfo{journal}{Journal of Engineering Mechanics} \bibinfo{year}{2021};\bibinfo{volume}{147}(\bibinfo{number}{11}):\bibinfo{pages}{04021082}.
\bibitem[{Kerschen and Golinval(2002)}]{kerschen2002physical}
\bibinfo{author}{Kerschen\xfnm[ G.]}, \bibinfo{author}{Golinval\xfnm[ J.C.]}.
\newblock \bibinfo{title}{Physical interpretation of the proper orthogonal modes using the singular value decomposition}.
\newblock \bibinfo{journal}{Journal of Sound and Vibration} \bibinfo{year}{2002};\bibinfo{volume}{249}(\bibinfo{number}{5}):\bibinfo{pages}{849--865}.
\bibitem[{Volkwein(2013)}]{volkwein2013proper}
\bibinfo{author}{Volkwein\xfnm[ S.]}.
\newblock \bibinfo{title}{Proper orthogonal decomposition: Theory and reduced-order modelling}.
\newblock \bibinfo{journal}{Lecture Notes, University of Konstanz} \bibinfo{year}{2013};\bibinfo{volume}{4}(\bibinfo{number}{4}):\bibinfo{pages}{1--29}.
\bibitem[{Zhang et~al.(2020)Zhang, Liu and Sun}]{zhang2020physics}
\bibinfo{author}{Zhang\xfnm[ R.]}, \bibinfo{author}{Liu\xfnm[ Y.]}, \bibinfo{author}{Sun\xfnm[ H.]}.
\newblock \bibinfo{title}{Physics-informed multi-{LSTM} networks for metamodeling of nonlinear structures}.
\newblock \bibinfo{journal}{Computer Methods in Applied Mechanics and Engineering} \bibinfo{year}{2020};\bibinfo{volume}{369}:\bibinfo{pages}{113226}.
\bibitem[{Li and Spence(2024)}]{li2024deep}
\bibinfo{author}{Li\xfnm[ B.]}, \bibinfo{author}{Spence\xfnm[ S.M.J.]}.
\newblock \bibinfo{title}{Deep learning enabled rapid nonlinear time history wind performance assessment}.
\newblock In: \bibinfo{booktitle}{Structures}; vol.~\bibinfo{volume}{66}. \bibinfo{organization}{Elsevier}; \bibinfo{year}{2024}, p. \bibinfo{pages}{106810}.
\bibitem[{Atila and Spence(2025)}]{atila2025metamodeling}
\bibinfo{author}{Atila\xfnm[ H.]}, \bibinfo{author}{Spence\xfnm[ S.M.J.]}.
\newblock \bibinfo{title}{Metamodeling of the response trajectories of nonlinear stochastic dynamic systems using physics-informed {LSTM} networks}.
\newblock \bibinfo{journal}{Journal of Building Engineering} \bibinfo{year}{2025};\bibinfo{volume}{111}:\bibinfo{pages}{113447}.
\bibitem[{Wu et~al.(2023)Wu, Yin, Zhang and Geng}]{wu2023prediction}
\bibinfo{author}{Wu\xfnm[ Y.]}, \bibinfo{author}{Yin\xfnm[ Z.]}, \bibinfo{author}{Zhang\xfnm[ H.]}, \bibinfo{author}{Geng\xfnm[ W.]}.
\newblock \bibinfo{title}{Prediction of nonlinear seismic response of underground structures in single- and multi-layered soil profiles using a deep gated recurrent network}.
\newblock \bibinfo{journal}{Soil Dynamics and Earthquake Engineering} \bibinfo{year}{2023};\bibinfo{volume}{168}:\bibinfo{pages}{107852}.
\bibitem[{Gao et~al.(2025)Gao, Peng, Xu, Guo and Chen}]{gao2025dynamic}
\bibinfo{author}{Gao\xfnm[ X.]}, \bibinfo{author}{Peng\xfnm[ C.]}, \bibinfo{author}{Xu\xfnm[ W.]}, \bibinfo{author}{Guo\xfnm[ T.]}, \bibinfo{author}{Chen\xfnm[ C.]}.
\newblock \bibinfo{title}{Dynamic time history response prediction through an experimentally trained deep gated recurrent units network using cyber-physical real-time hybrid simulation}.
\newblock \bibinfo{journal}{Mechanical Systems and Signal Processing} \bibinfo{year}{2025};\bibinfo{volume}{224}:\bibinfo{pages}{112247}.
\bibitem[{Shewalkar et~al.(2019)Shewalkar, Nyavanandi and Ludwig}]{shewalkar2019performance}
\bibinfo{author}{Shewalkar\xfnm[ A.]}, \bibinfo{author}{Nyavanandi\xfnm[ D.]}, \bibinfo{author}{Ludwig\xfnm[ S.A.]}.
\newblock \bibinfo{title}{Performance evaluation of deep neural networks applied to speech recognition: {RNN}, {LSTM}, and {GRU}}.
\newblock \bibinfo{journal}{Journal of Artificial Intelligence and Soft Computing Research} \bibinfo{year}{2019};\bibinfo{volume}{9}(\bibinfo{number}{4}):\bibinfo{pages}{235--245}.
\bibitem[{Nosouhian et~al.(2021)Nosouhian, Nosouhian and Kazemi~Khoshouei}]{nosouhian2021review}
\bibinfo{author}{Nosouhian\xfnm[ S.]}, \bibinfo{author}{Nosouhian\xfnm[ F.]}, \bibinfo{author}{Kazemi~Khoshouei\xfnm[ A.]}.
\newblock \bibinfo{title}{A review of recurrent neural network architecture for sequence learning: Comparison between {LSTM} and {GRU}}.
\newblock \bibinfo{journal}{Preprints} \bibinfo{year}{2021};.
\bibitem[{Srivastava et~al.(2014)Srivastava, Hinton, Krizhevsky, Sutskever and Salakhutdinov}]{srivastava2014dropout}
\bibinfo{author}{Srivastava\xfnm[ N.]}, \bibinfo{author}{Hinton\xfnm[ G.]}, \bibinfo{author}{Krizhevsky\xfnm[ A.]}, \bibinfo{author}{Sutskever\xfnm[ I.]}, \bibinfo{author}{Salakhutdinov\xfnm[ R.]}.
\newblock \bibinfo{title}{Dropout: A simple way to prevent neural networks from overfitting}.
\newblock \bibinfo{journal}{The Journal of Machine Learning Research} \bibinfo{year}{2014};\bibinfo{volume}{15}(\bibinfo{number}{1}):\bibinfo{pages}{1929--1958}.
\bibitem[{Cohen et~al.(1992)Cohen, Daubechies and Feauveau}]{cohen1992biorthogonal}
\bibinfo{author}{Cohen\xfnm[ A.]}, \bibinfo{author}{Daubechies\xfnm[ I.]}, \bibinfo{author}{Feauveau\xfnm[ J.C.]}.
\newblock \bibinfo{title}{Biorthogonal bases of compactly supported wavelets}.
\newblock \bibinfo{journal}{Communications on Pure and Applied Mathematics} \bibinfo{year}{1992};\bibinfo{volume}{45}(\bibinfo{number}{5}):\bibinfo{pages}{485--560}.
\bibitem[{Le and Caracoglia(2015)}]{le2015reduced}
\bibinfo{author}{Le\xfnm[ T.H.]}, \bibinfo{author}{Caracoglia\xfnm[ L.]}.
\newblock \bibinfo{title}{Reduced-order wavelet-galerkin solution for the coupled, nonlinear stochastic response of slender buildings in transient winds}.
\newblock \bibinfo{journal}{Journal of Sound and Vibration} \bibinfo{year}{2015};\bibinfo{volume}{344}:\bibinfo{pages}{179--208}.
\bibitem[{Wang and Wu(2020)}]{wang2020knowledge}
\bibinfo{author}{Wang\xfnm[ H.]}, \bibinfo{author}{Wu\xfnm[ T.]}.
\newblock \bibinfo{title}{Knowledge-enhanced deep learning for wind-induced nonlinear structural dynamic analysis}.
\newblock \bibinfo{journal}{Journal of Structural Engineering} \bibinfo{year}{2020};\bibinfo{volume}{146}(\bibinfo{number}{11}):\bibinfo{pages}{04020235}.
\bibitem[{Fushiki(2011)}]{fushiki2011estimation}
\bibinfo{author}{Fushiki\xfnm[ T.]}.
\newblock \bibinfo{title}{Estimation of prediction error by using {K}-fold cross-validation}.
\newblock \bibinfo{journal}{Statistics and Computing} \bibinfo{year}{2011};\bibinfo{volume}{21}:\bibinfo{pages}{137--146}.
\bibitem[{Peherstorfer(2019)}]{peherstorfer2019multifidelity}
\bibinfo{author}{Peherstorfer\xfnm[ B.]}.
\newblock \bibinfo{title}{Multifidelity {M}onte {C}arlo estimation with adaptive low-fidelity models}.
\newblock \bibinfo{journal}{SIAM/ASA Journal on Uncertainty Quantification} \bibinfo{year}{2019};\bibinfo{volume}{7}(\bibinfo{number}{2}):\bibinfo{pages}{579--603}.
\bibitem[{Chuang and Spence(2019)}]{chuang2019efficient}
\bibinfo{author}{Chuang\xfnm[ W.C.]}, \bibinfo{author}{Spence\xfnm[ S.M.J.]}.
\newblock \bibinfo{title}{An efficient framework for the inelastic performance assessment of structural systems subject to stochastic wind loads}.
\newblock \bibinfo{journal}{Engineering Structures} \bibinfo{year}{2019};\bibinfo{volume}{179}:\bibinfo{pages}{92--105}.
\bibitem[{Duarte et~al.(2023)Duarte, Arunachalam, Subgranon and Spence}]{Duarte2023}
\bibinfo{author}{Duarte\xfnm[ T.G.A.]}, \bibinfo{author}{Arunachalam\xfnm[ S.]}, \bibinfo{author}{Subgranon\xfnm[ A.]}, \bibinfo{author}{Spence\xfnm[ S.M.J.]}.
\newblock \bibinfo{title}{Uncertainty quantification and simulation of wind-tunnel-informed stochastic wind loads}.
\newblock \bibinfo{journal}{Wind} \bibinfo{year}{2023};\bibinfo{volume}{3}(\bibinfo{number}{3}):\bibinfo{pages}{375--393}.
\bibitem[{{Tokyo Polytechnic University (TPU)}(2007)}]{TPU2007winddb}
\bibinfo{author}{{Tokyo Polytechnic University (TPU)}\xfnm[]}.
\newblock \bibinfo{title}{{TPU} aerodynamic wind tunnel database}.
\newblock \bibinfo{howpublished}{Wind Engineering Information Center, Tokyo Polytechnic University}; \bibinfo{year}{2007}.
\newblock \URLprefix \url{https://db.wind.arch.t-kougei.ac.jp/}; \bibinfo{note}{low- and high-rise building pressure data; accessed \today}.
\bibitem[{Mazzoni et~al.(2006)Mazzoni, McKenna, Scott, Fenves et~al.}]{mazzoni2006opensees}
\bibinfo{author}{Mazzoni\xfnm[ S.]}, \bibinfo{author}{McKenna\xfnm[ F.]}, \bibinfo{author}{Scott\xfnm[ M.H.]}, \bibinfo{author}{Fenves\xfnm[ G.L.]}, et~al.
\newblock \bibinfo{title}{{OpenSees} command language manual}.
\newblock \bibinfo{journal}{Pacific Earthquake Engineering Research (PEER) Center} \bibinfo{year}{2006};\bibinfo{volume}{264}(\bibinfo{number}{1}):\bibinfo{pages}{137--158}.
\bibitem[{Filippou et~al.(1983)Filippou, Popov and Bertero}]{filippou1983effects}
\bibinfo{author}{Filippou\xfnm[ F.C.]}, \bibinfo{author}{Popov\xfnm[ E.P.]}, \bibinfo{author}{Bertero\xfnm[ V.V.]}.
\newblock \bibinfo{title}{Effects of bond deterioration on hysteretic behavior of reinforced concrete joints}.
\newblock \bibinfo{journal}{Earthquake Engineering Research Center, University of California, Berkeley} \bibinfo{year}{1983};.
\bibitem[{Ballio and Castiglioni(1995)}]{ballio1995unified}
\bibinfo{author}{Ballio\xfnm[ G.]}, \bibinfo{author}{Castiglioni\xfnm[ C.A.]}.
\newblock \bibinfo{title}{A unified approach for the design of steel structures under low and/or high cycle fatigue}.
\newblock \bibinfo{journal}{Journal of Constructional Steel Research} \bibinfo{year}{1995};\bibinfo{volume}{34}(\bibinfo{number}{1}):\bibinfo{pages}{75--101}.
\bibitem[{Li et~al.(2023)Li, Chuang and Spence}]{li2023reliability}
\bibinfo{author}{Li\xfnm[ B.]}, \bibinfo{author}{Chuang\xfnm[ W.C.]}, \bibinfo{author}{Spence\xfnm[ S.M.J.]}.
\newblock \bibinfo{title}{Reliability of inelastic wind-excited structures by dynamic shakedown and adaptive fast nonlinear analysis ({AFNA})}.
\newblock \bibinfo{journal}{Engineering Structures} \bibinfo{year}{2023};\bibinfo{volume}{296}:\bibinfo{pages}{116869}.
\bibitem[{Xu and Spence(2025)}]{xu2025multiple}
\bibinfo{author}{Xu\xfnm[ L.]}, \bibinfo{author}{Spence\xfnm[ S.M.J.]}.
\newblock \bibinfo{title}{Multiple stripe analysis for rapid failure probability analysis in support of performance-based wind engineering}.
\newblock \bibinfo{journal}{Engineering Structures} \bibinfo{year}{2025};\bibinfo{volume}{342}:\bibinfo{pages}{120864}.
\bibitem[{Spence and Kareem(2013)}]{spence2013data}
\bibinfo{author}{Spence\xfnm[ S.M.J.]}, \bibinfo{author}{Kareem\xfnm[ A.]}.
\newblock \bibinfo{title}{Data-enabled design and optimization ({DEDOpt}): Tall steel building frameworks}.
\newblock \bibinfo{journal}{Computers \& Structures} \bibinfo{year}{2013};\bibinfo{volume}{129}:\bibinfo{pages}{134--147}.

\end{thebibliography}

\end{document}